\documentclass[review]{bmvc2k}

\usepackage{array,multirow,rotating,siunitx}
\usepackage{booktabs}
\usepackage{pifont}
\usepackage{amssymb}
\usepackage{amsfonts}
\usepackage{comment}
\usepackage{array}
\newcolumntype{C}[1]{>{\centering\arraybackslash}p{#1}}

\usepackage{color, colortbl}
\usepackage{array}
\definecolor{mygray}{gray}{0.91}

\newif\ifcomments
\commentstrue 
\ifcomments
	\newcommand{\carlo}[1]{\textcolor{red}{[#1]}}
	\newcommand{\shuzhi}[1]{\textcolor{blue}{[#1]}}
    \newcommand{\hannah}[1]{\textcolor{magenta}{#1}}
\else
	\newcommand{\carlo}[1]{}
	\newcommand{\shuzhi}[1]{}
    \newcommand{\hannah}[1]{}
\fi

\newcommand{\bd}{\mathbf{d}}
\newcommand{\bff}{\mathbf{f}}

\newcommand{\bx}{\mathbf{x}}

\newcommand{\tightparagraph}{\vspace*{-0.02\textwidth}\paragraph}

\newcommand{\ie}{\textit{i.e.}}

\title{Joint Detection of Motion Boundaries and Occlusions}
\addauthor{Hannah Halin Kim}{hannah@cs.duke.edu}{1}
\addauthor{Shuzhi Yu}{shuzhiyu@cs.duke.edu}{1}
\addauthor{Carlo Tomasi}{tomasi@cs.duke.edu}{1}

\addinstitution{
 Duke University\\
 Durham, NC, USA
}

\runninghead{Kim, Yu, Tomasi}{Joint Detection of Motion Boundaries and Occlusions}

\def\etal{\emph{et al}\bmvaOneDot}

\begin{document}
\maketitle

\begin{abstract}
We propose MONet, a convolutional neural network that jointly detects motion boundaries (MBs) and occlusion regions (Occs) in video both forward and backward in time. Detection is difficult because optical flow is discontinuous along MBs and undefined in Occs, while many flow estimators assume smoothness and a flow defined everywhere. To reason in the two time directions simultaneously, we direct-warp the estimated maps between the two frames. Since appearance mismatches between frames often signal vicinity to MBs or Occs, we construct a cost block that for each feature in one frame records the lowest discrepancy with matching features in a search range. This cost block is two-dimensional, and much less expensive than the four-dimensional cost volumes used in flow analysis. Cost-block features are computed by an encoder, and MB and Occ estimates are computed by a decoder. We found that arranging decoder layers fine-to-coarse, rather than coarse-to-fine, improves performance. MONet outperforms the prior state of the art for both tasks on the Sintel and FlyingChairsOcc benchmarks without any fine-tuning on them.
\end{abstract}

\section{Introduction}
Thanks to large-scale video datasets~\cite{mpi,flythings3D,flownet} and advances in deep learning, recent work~\cite{PWC,refine_of_occ,raft} has rapidly improved dense optical flow estimation via learning with Convolutional Neural Network (CNNs). However, flow predictors still suffer near \textit{motion boundaries} (MBs), the curves across which the optical flow field is discontinuous~\cite{LDMB}, and in \textit{occlusion regions} (Occs), sets of pixels in one frame that do not have correspondences in the other. First, flow estimates are typically regularized by imposing spatial smoothness, which harms predictions near MBs. Second, flow cannot be measured from the input images in Occs and can only be plausibly guessed from the statistics of the ground truth provided in synthetic datasets. For instance, the top flow estimator RAFT~\cite{raft} achieves an End-Point Error of 1.4 pixels on Sintel~\cite{mpi}, but of 6.5 for pixels that are within 5 pixels from a MB and 4.7 in Occs. The accurate detection of MBs and Occs helps understand where flow estimates can be trusted and provides important visual cues for tracking~\cite{goodfeatures} and video segmentation~\cite{mb_vos}.

\begin{figure}[!t]
    \begin{center}
        \setlength{\tabcolsep}{6pt}
        \begin{tabular}{cc}
            \begin{tabular}[b]{cc}
                \raisebox{5.3mm}{Frame 1} & \includegraphics[width=0.224\textwidth]{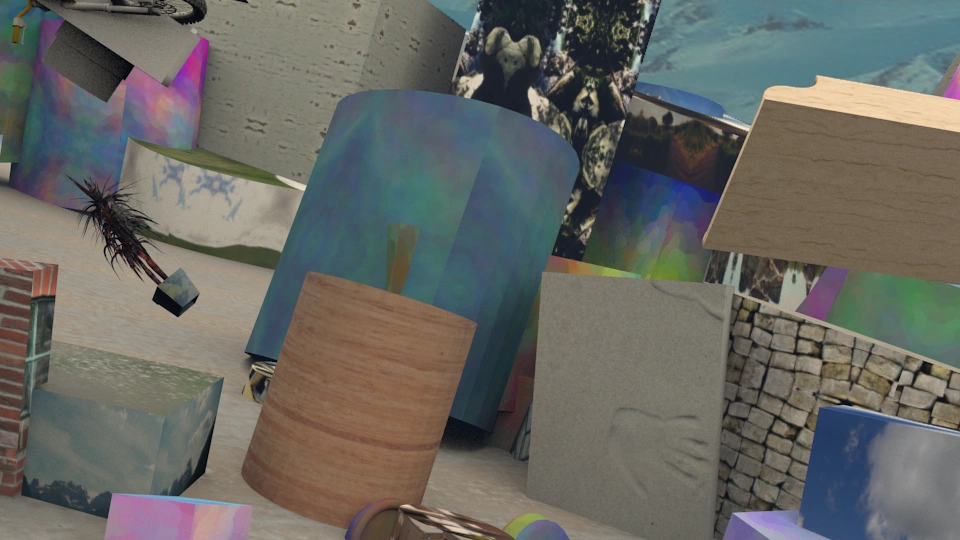} \\
                \raisebox{5.3mm}{Frame 2} & \includegraphics[width=0.224\textwidth]{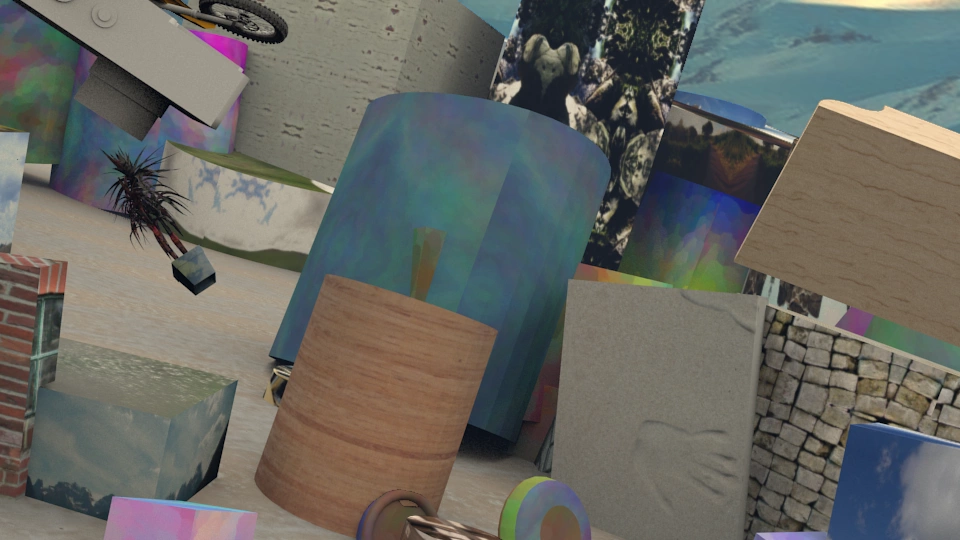}
            \end{tabular} &
            \begin{tabular}[b]{c}
                \includegraphics[width=0.47\textwidth]{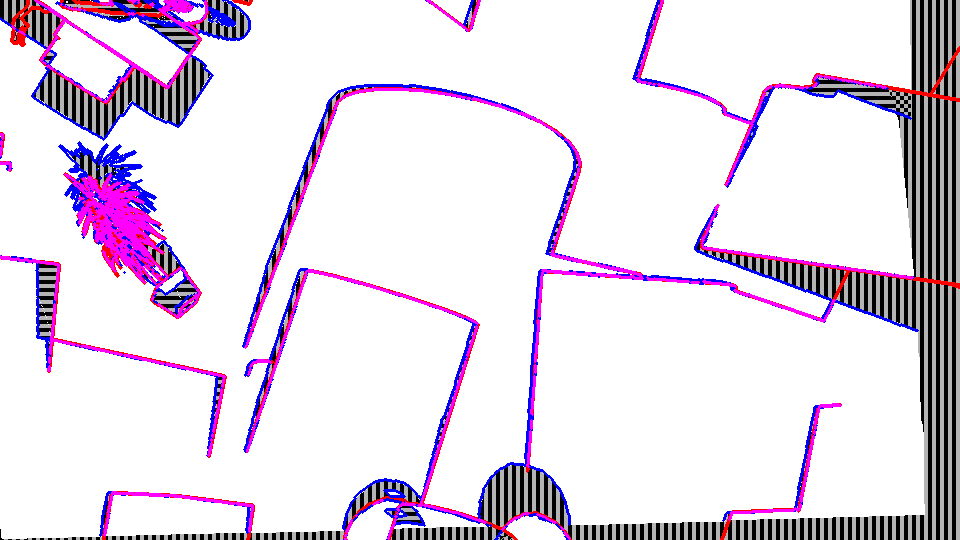}
            \end{tabular}
        \end{tabular}
    \end{center}
    \caption{Occs align with MBs in this FlyingThings3D~\cite{flythings3D} frame pair. Vertical stripes denote $O_1$, \ie, Occs in frame 1. Horizontal stripes are direct-warps to frame 1 of Occs in frame 2 ($D(O_2)$), and checkerboard patterns are $O_1 \cap D(O_2)$. Red curves are $M_1$ (similar notation for MBs $M$ as for Occs $O$), blue ones are $D(M_2)$, and purple ones are $M_1 \cap D(M_2)$.
    }
    \label{fig:occ_grad_mb_align}
\end{figure}

Occs occur near MBs, where motion is discontinuous (Figure \ref{fig:occ_grad_mb_align}). 
In addition, disocclusions in one time direction are occlusions in the other. This suggests estimating MBs and Occs jointly, and to reason in both time directions simultaneously. Accordingly, we propose a CNN named \textit{MONet} to jointly estimate MBs and Occs given two consecutive images and their estimated flow~\cite{PWC}. The network uses Siamese networks to leverage time symmetry.

Simultaneous reasoning in the two time directions requires mapping all quantities between frames, and we do this in a novel way. All previous flow~\cite{raft, PWC, flownet}, Occ~\cite{refine_of_occ}, and MB~\cite{eccv18_mb} estimators use the flow from frame $b$ to $a$ to \textit{reverse warp} features from frame $a$ to frame $b$. We instead use flow from frame $a$ to $b$ to \textit{direct warp} features from frame $a$ to frame $b$. We show that direct warping both preserves features in Occs and provides additional Occ information through the regions it leaves undefined (Figure \ref{fig:mb_occ_ex}).

MBs often contour Occs. Based on this observation, we propose to make the MB predictor of MONet focus on Occ boundaries. Specifically, we use an attention mechanism~\cite{Zhang_2018_CVPR} to place MB predictions where the gradient magnitude of the Occ map is large, that is, along predicted Occ boundaries. We use the MB labels to further supervise the attention map.

Since both MBs and Occs disrupt correspondences, MONet computes \textit{cost blocks} that measure the lowest discrepancy between each feature of the first frame and its matching features in a search window in the second frame. A cost block is two-dimensional, and is the minimum over the search window of the four-dimensional cost volumes used in previous estimators of flow~\cite{raft, PWC, flownet}, MBs~\cite{eccv18_mb}, and Occs~\cite{refine_of_occ}. The light-weight cost blocks are sufficient for our purposes and much less expensive to work with than cost volumes.

\begin{figure}[!t]
    \centering
    \begin{tabular}{cc} 
    \includegraphics[width=5cm]{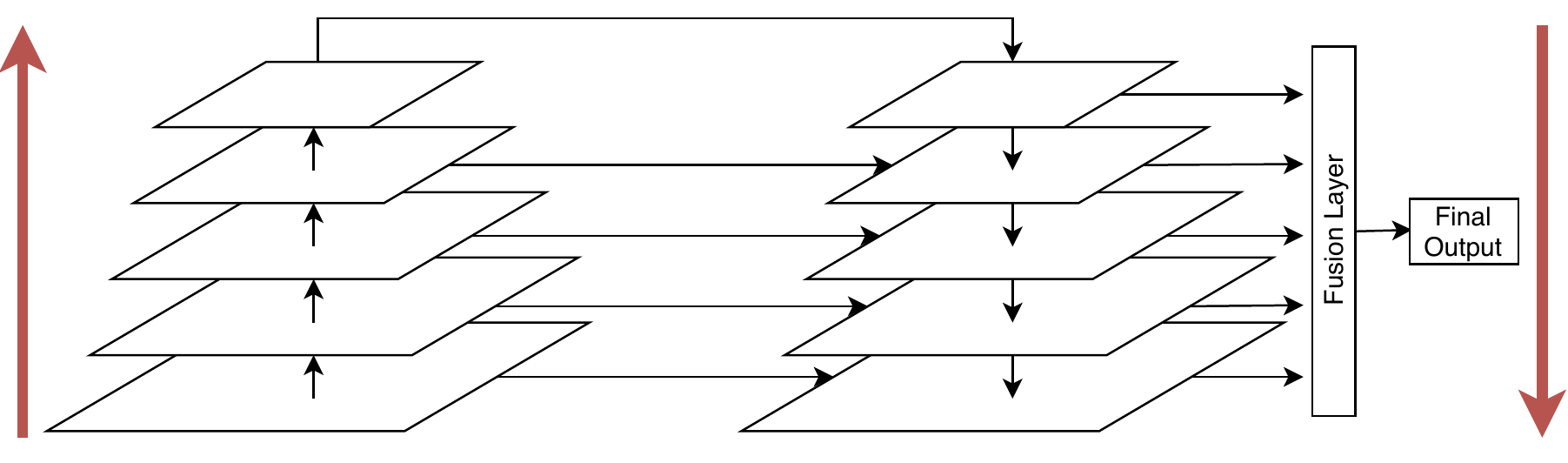} &
    \includegraphics[width=5cm]{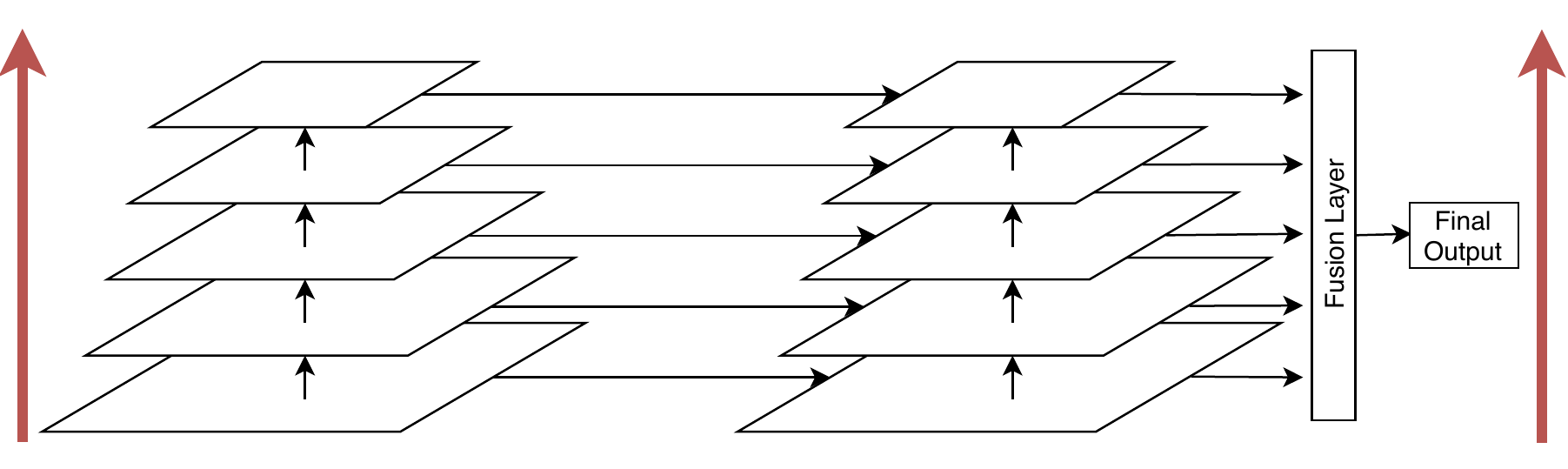}\\ 
    (a) & (b) \\ 
    \end{tabular}
       \caption{(a) An encoder-decoder network~\cite{fcn}. (b) Our architecture with fine-to-coarse predictor (b). The encoder is the same, but information flow in the decoder is reversed. \label{fig:cnn_architectures}}
\end{figure}

Cost blocks are defined on features computed by the MONet encoder, and a two-branch decoder then computes MB and Occ predictions. While the decoders in all previous estimators that use cost volumes~\cite{flownet,refine_of_occ,cv_of_occ,PWC,eccv18_mb} process information from coarse to fine, we do so Fine-to-Coarse (F2C) and show empirically the benefits of doing so (see Figure \ref{fig:cnn_architectures}). 

\setlength{\tabcolsep}{15pt} 

\tightparagraph{Summary of Contributions:} 
Direct warping to better preserve information between frames; Spatial attention mechanism to align MB and Occ predictions; Two-dimensional cost blocks to measure feature discrepancies between frames; Fine-to-coarse decoder for higher accuracy; State-of-the-art performance for both MB and Occ detection on Sintel and FlyingChairsOcc without any fine-tuning on either dataset, and even after several ablations.

\section{Related Work} \label{relatedworks}
In spite of continued advances~\cite{fullflow,flownet,of0,PWC,dcflow}, accurate \textbf{optical flow} estimation remains an open challenge especially near MBs and Occs. Recent methods~\cite{PWC,dcflow} starting with Chen and Koltun~\cite{fullflow} use a four-dimensional \textbf{cost volumes}~\cite{cv} to compute optical flow. DC Flow~\cite{dcflow} and FlowNet~\cite{flownet} build the full cost volume at a single scale, and Sun \etal~\cite{PWC} show that building cost volumes at multiple scales leads to better models. Teed and Deng~\cite{raft} achieve SOTA performance with RAFT, a deep network that builds a complete multi-scale four-dimensional cost volumes for all pairs of pixels on the input images. MONet instead uses a simplified two-dimensional cost block for MB and Occ detection. 

The related task of detecting \textbf{occlusion regions} has recently received considerable attention~\cite{occ4,occ2,refine_of_occ,cv_of_occ}. One Occ detector~\cite{occ2} takes as input optical flow estimated with four different algorithms and trains random forests to classify pixels into Occ or non-Occ categories. Fu \etal~\cite{occ4} use CNNs to detect Occ boundaries at the patch level and connect these detections to each other with a conditional random field~\cite{crf}. Hur and Roth~\cite{refine_of_occ} achieve SOTA performance using a CNN to infer Occs and flow jointly. Neoral \etal~\cite{cv_of_occ} consider the two problems sequentially by first detecting Occs and then using Occs to help estimate flow. MONet jointly solves the two closely related problem of MB and Occ detection.

Early work~\cite{spoerri} on \textbf{motion boundary} estimation exploits the observation that local flow histograms are bimodal near MBs. Liu \etal~\cite{mb0} propose to detect MBs by tracking and grouping hypothetical motion edge fragments bottom-up in scale. LDMB~\cite{LDMB} uses structured random forests~\cite{sed} and takes as inputs two consecutive images, optical flow estimates~\cite{classicnl}, and image warping errors, but produces noisy boundaries and fails on small and thin objects. In addition to the forward flow estimation from frame $i$ to $i+1$, LDMB also takes as input the backward flow estimation from frame $i$ to $i-1$ and requires three frames. MO-Net instead utilizes bi-directional flow between frames $i$ and $i+1$, and only require two frames. Ilg \etal~\cite{eccv18_mb} uses CNNs incorporating joint training and refinement to simultaneously estimate Occs, MBs, optical flow, disparities, motion segmentation, and scene flow in both temporal directions. However, they simply stack multiple networks for joint-task solving and do not explicitly utilize the relationships between these tasks. MBs have also been used to aid in other low-level vision tasks such as video object segmentation~\cite{mb_vos}.

Kramer's auto-encoder~\cite{kramer1991nonlinear} is a precursor to the \textbf{encoder-decoder architecture} with three hidden layers with an information bottleneck. Long \textit{et al.} proposes a Fully Convolutional Network, a fully-fledged encoder-decoder.  
Subsequent work makes upsampling deeper~\cite{deconvnet} and symmetrizes the architecture into what is called a U-Net~\cite{ronneberger2015u}.
The paper by Schulz and Behnke~\cite{schulz2012learning} proposes a shallow encoder followed by an F2C decoder, and uses only the coarsest prediction for inference. Our model is deeper and employs a trainable fusion layer, as we observed that the coarsest prediction is not always the best. 

\section{Principles} \label{proposed}

\begin{figure}[!]
    \centering
    \includegraphics[width=\textwidth]{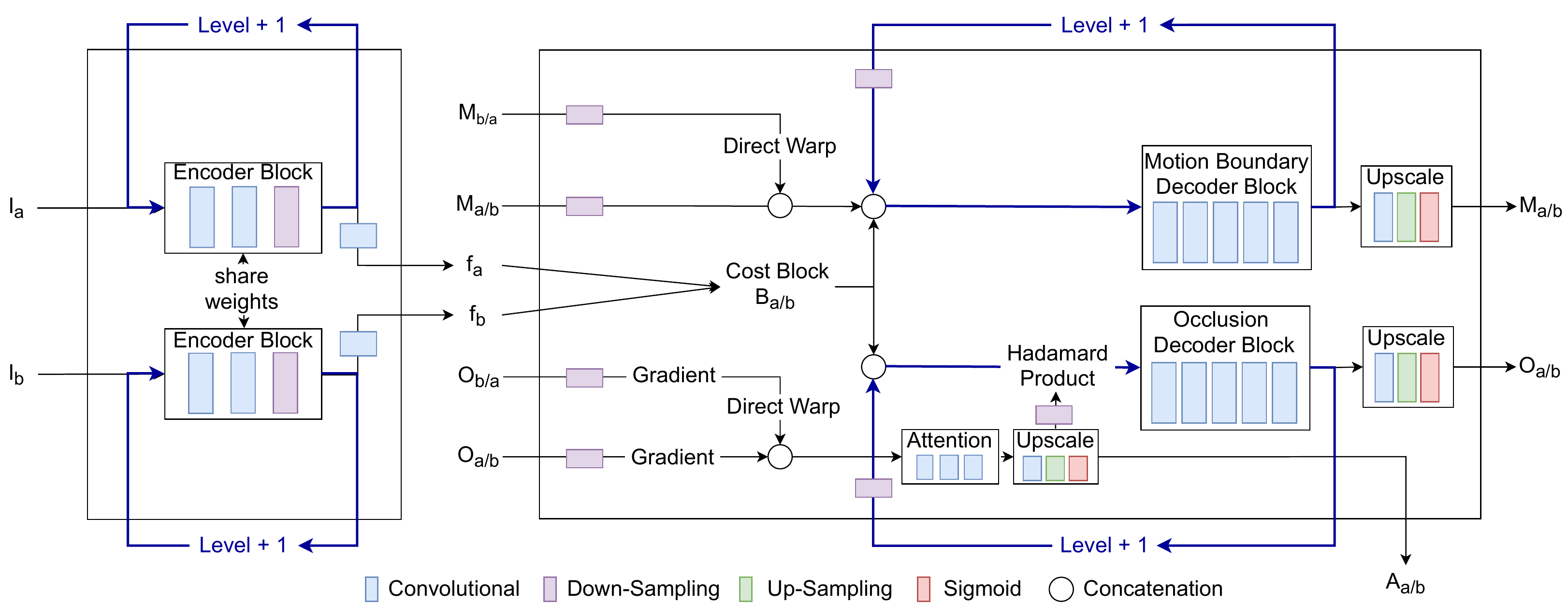}
    \caption{MONet predicts bidirectional $M_{a/b}$ and $O_{a/b}$
    maps in each level $l$ using bi-directional predictions ($M_a$, $M_b$, $O_a$, $O_b$) and cost block ($B_{a/b}$) from the previous level $l-1$. 
    Blue arrows represent the main flow of information across levels.}
    \label{fig:mb_occ_level}
\end{figure}

Section \ref{sec:architecture} describes MONet, a new neural network that jointly predicts MB scores $M \in [0, 1]^{w \times h}$ and Occ scores $O \in [0, 1]^{w \times h}$ both forward and backward in time (Figure \ref{fig:mb_occ_level}). It takes as input two consecutive video frames $I_1, I_2 \in \mathbb{R}^{w \times h \times 3}$ and the corresponding bi-directional flow estimates $F \in \mathbb{R}^{w \times h \times 2}$ from an existing flow estimator. This Section describes the new principles that MONet embodies.

A first idea is that analysis of image motion in one time direction supports analysis in the other. As a result, the same inputs are provided in both temporal orders, $1\rightarrow 2$ and $2\rightarrow 1$. The model's parametric complexity is kept constant by sharing weights between temporal directions. Second, Occs align with MBs, and predicting Occs and MBs jointly yields richer insights than separate analyses would. This suggests a network with one Siamese encoder branch and two Siamese decoder branches that exchange information at all levels.

These principles are supported by the following technical ideas, described next: (i) Direct warping is more useful than reverse warping when aligning maps across frames; (ii) Attention can help align MBs and Occs; (iii) Inexpensive cost blocks capture useful information for MB and Occ detection. Architectural considerations are left for Section \ref{sec:architecture}.

\newcommand{\plabel}[1]{\textit{#1}}
\begin{figure}[!t]
    \newcommand{\warppanel}[1]{
        \includegraphics[width=0.194\textwidth]{images/warping/warp_#1.pdf}
    }
    \begin{center}
        \setlength{\tabcolsep}{0pt}
        \begin{tabular}{*{5}{c}}
            \warppanel{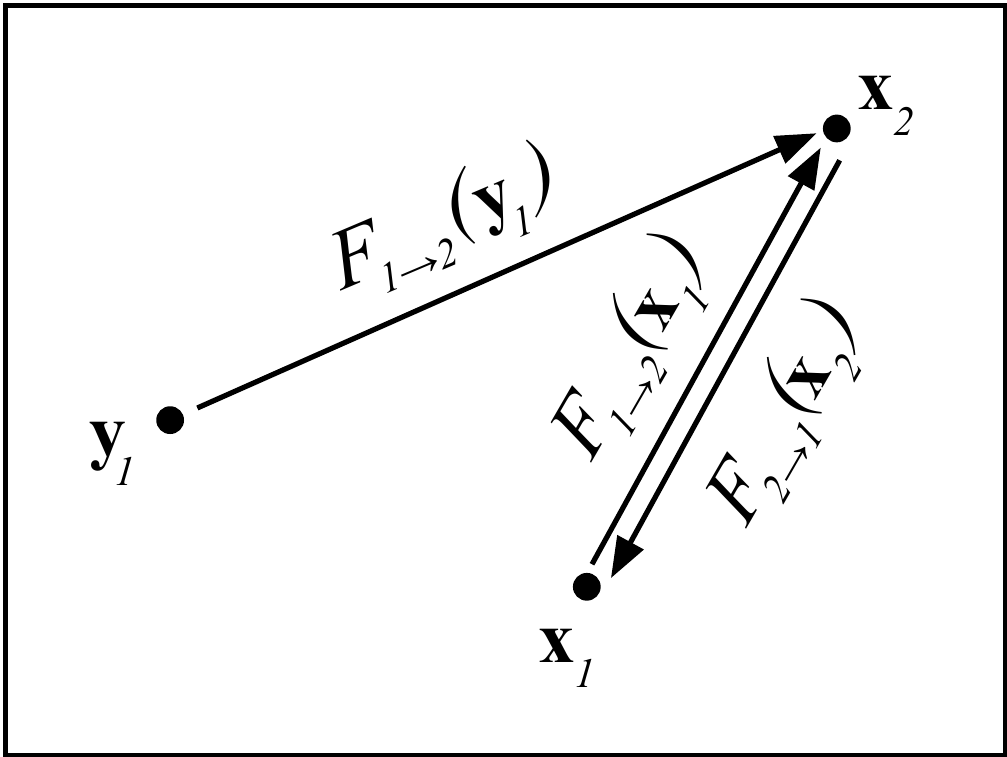} &
            \warppanel{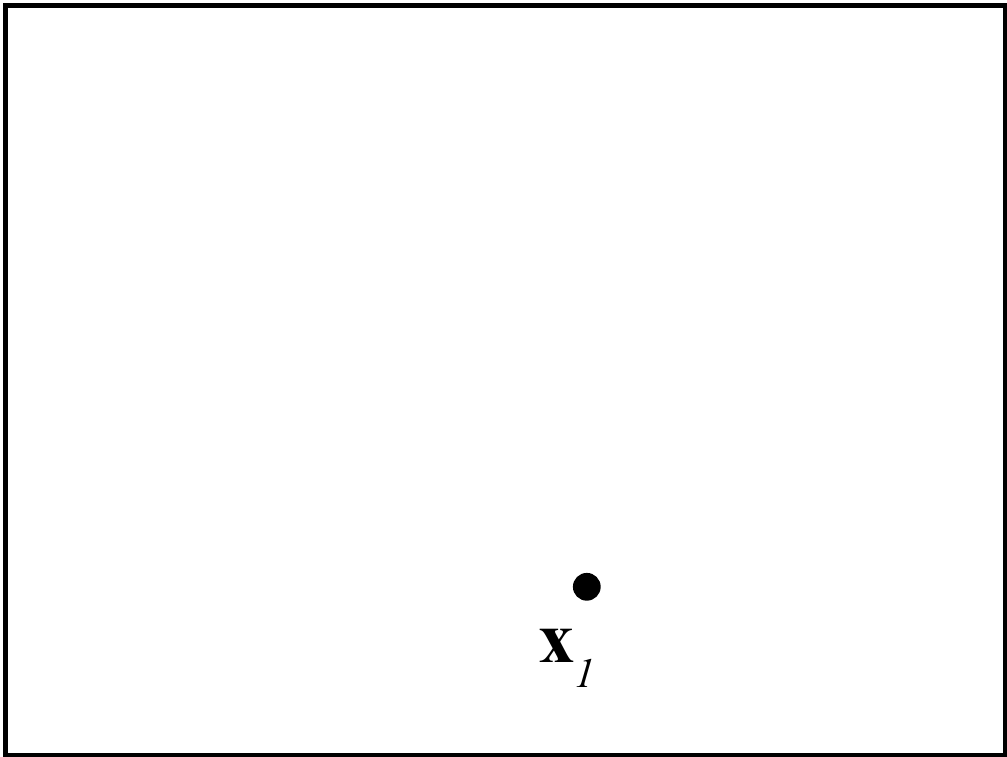} &
            \warppanel{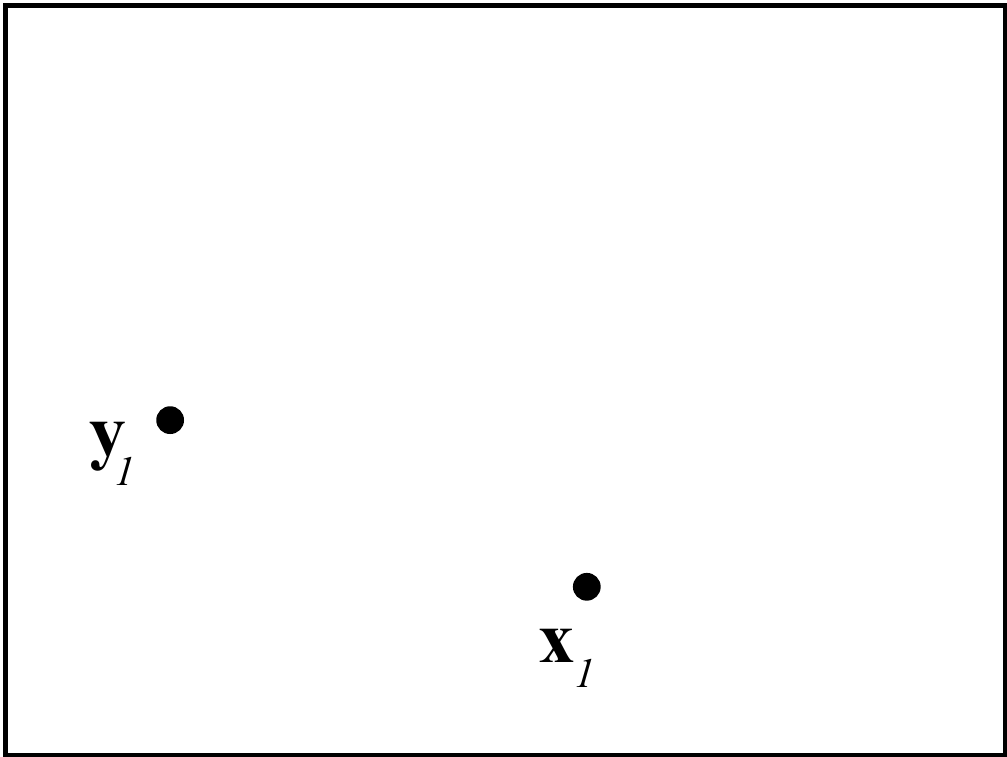} &
            \warppanel{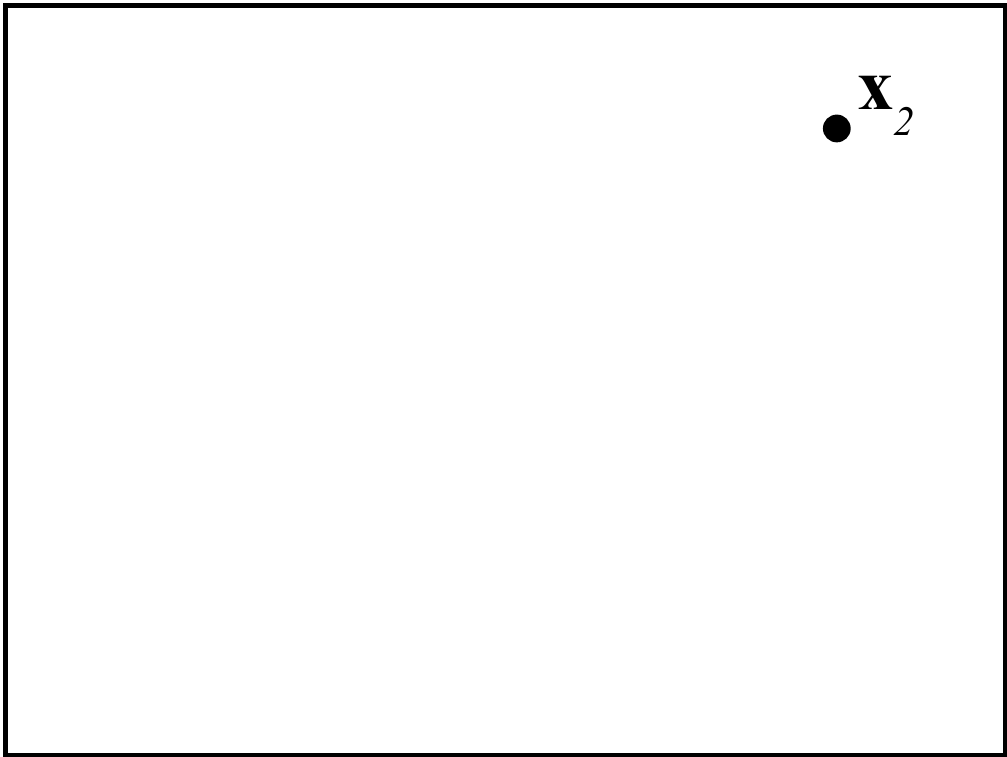} &
            \warppanel{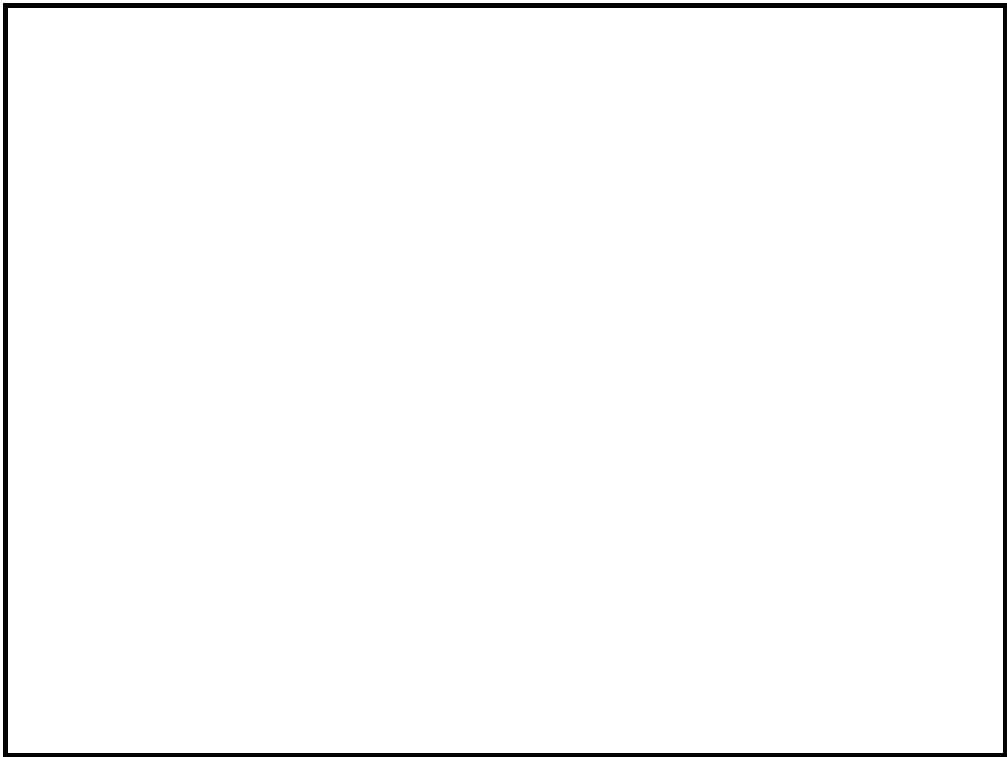} \\
            \plabel{a}: flows &
            \plabel{b}: $\text{direct}(\mathbf{x}_2)$ &
            \plabel{c}: $\text{reverse}(\mathbf{x}_2)$ &
            \plabel{d}: $\text{direct}(\mathbf{y}_1)$ &
            \plabel{e}: $\text{reverse}(\mathbf{y}_1)$
        \end{tabular}
    \end{center}
    \caption{Direct and reverse warping in the presence of occlusion. \plabel{(a)} Points $x$ and $y$ at $\mathbf{x_1}$ and $\mathbf{y_1}$ in frame 1 move to the same point $\mathbf{x_2}$ in frame 2 and $y$ becomes occluded behind $x$ at $\mathbf{x_2}$. If flow is modeled as a function, there is only one flow $F_{2\rightarrow 1}(\mathbf{x}_2)$ in the $2\rightarrow 1$ direction. \plabel{(b)} Direct warping of $\mathbf{x_2}$ to frame 1 using $F_{2\rightarrow 1}$; \plabel{(c)} Reverse warping of $\mathbf{x_2}$ to frame 1 using $F_{1\rightarrow 2}$; \plabel{(d)} Direct warping of $\mathbf{y_1}$ to frame 2 using $F_{1\rightarrow2}$; \plabel{(e)} Reverse warping of $\mathbf{y_1}$ to frame 2 using  $F_{2\rightarrow 1}$. Since the flow from $\mathbf{x}_2$ to $\mathbf{y}_1$ is not defined, nothing gets mapped in this case.}
    \label{fig:fwd_bwd_warp_ex}
\end{figure}

\subsection{Direct Warping Provides Rich Occlusion Information}

Computing MB maps and Occ maps in both frames requires establishing inter-frame correspondences. If a point that is away from both MBs and Occs is at $\mathbf{x}_a$ in frame $a$ and at $\mathbf{x}_b$ in frame $b$, both flows are defined and unique:
\begin{equation}
\mathbf{x}_b = \mathbf{x}_a + F_{a\rightarrow b}(\mathbf{x}_a)
\;\;\;\text{and}\;\;\;
\mathbf{x}_a = \mathbf{x}_b + F_{b\rightarrow a}(\mathbf{x}_b)
\;\;\;\text{so that}\;\;\;
F_{a\rightarrow b}(\mathbf{x}_a) = -F_{b\rightarrow a}(\mathbf{x}_b) \;.
\end{equation}

To map image values $I_a$ from frame $a$ to estimated values $\hat{I}_b$ in frame $b$, one can then use \emph{direct warping}, which uses the flow defined in the same temporal direction as the map itself:
\begin{equation}
\hat{I}_b(\mathbf{x}_a + F_{a\rightarrow b}(\mathbf{x}_a)) = I_a(\mathbf{x}_a)
\label{eq:direct_warp}
\end{equation}
or \emph{reverse warping}, which uses the flow in the direction opposite to the desired map:
\begin{equation}
\hat{I}_b(\mathbf{x}_b) = I_a(\mathbf{x}_b + F_{b\rightarrow a}(\mathbf{x}_b))\;.
\label{eq:reverse_warp}
\end{equation}
Reverse warping is preferred when one loops over a grid of locations $\mathbf{x}_b$ to fill $\hat{I}_b$ completely. Pixel discretization aside, however, the result is the same either way.

The situation is more complex for points in Occs or on MBs. For example, Figure \ref{fig:fwd_bwd_warp_ex} (a) shows two points $x$ and $y$ at locations $\mathbf{x_1}$ and $\mathbf{y_1}$ in the first frame that both move to the same location $\mathbf{x_2}$ in the second frame. Point $x$ remains visible, while point $y$ becomes occluded behind point $x$. The map in the $2\rightarrow 1$ direction is one-to-many (not a function). However, it is customary (although somewhat arbitrary) to model flow as a function. If we do so, the flow at $\mathbf{x}_2$ in frame 2 maps to $\mathbf{x}_1$ in frame 1, and there is no flow from $\mathbf{x}_2$ to $\mathbf{y}_1$. The effects of direct and reverse mapping in the two directions are illustrated in the remaining panels.

\begin{figure}[!]
    \newcommand{\wimpanel}[1]{
        \includegraphics[width=0.117\textwidth]{images/warping/warp_img_#1.pdf}
    }
    \begin{center}
        \setlength{\tabcolsep}{0pt}
        \begin{tabular}{*{8}{c}}
            \multicolumn{3}{c}{$\overbrace{\hspace{0.36\textwidth}}^{\mbox{defined in $I_1$}}$} &
            \multicolumn{5}{c}{$\overbrace{\hspace{0.60\textwidth}}^{\mbox{defined in $I_2$}}$} \\
            \wimpanel{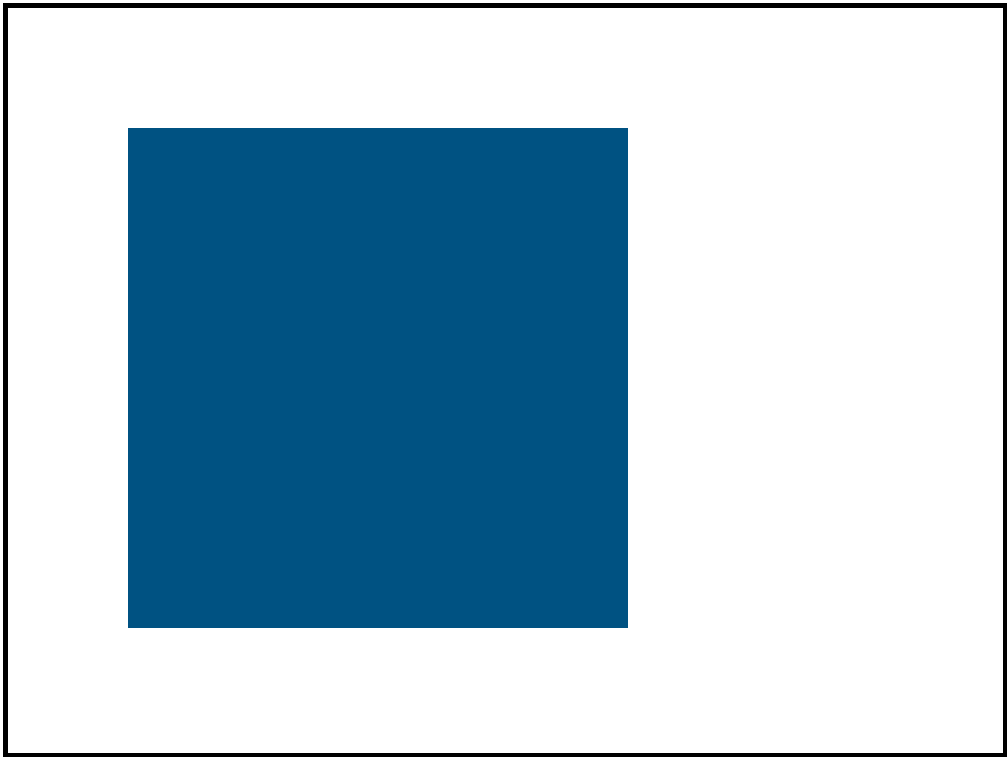} &
            \wimpanel{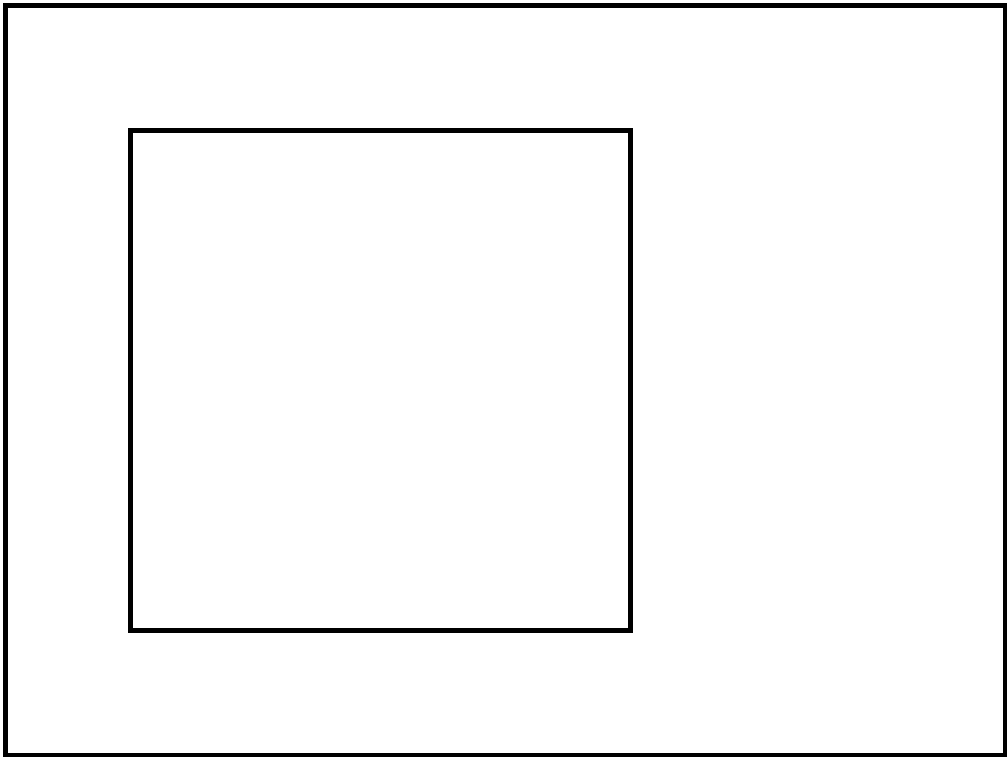} &
            \wimpanel{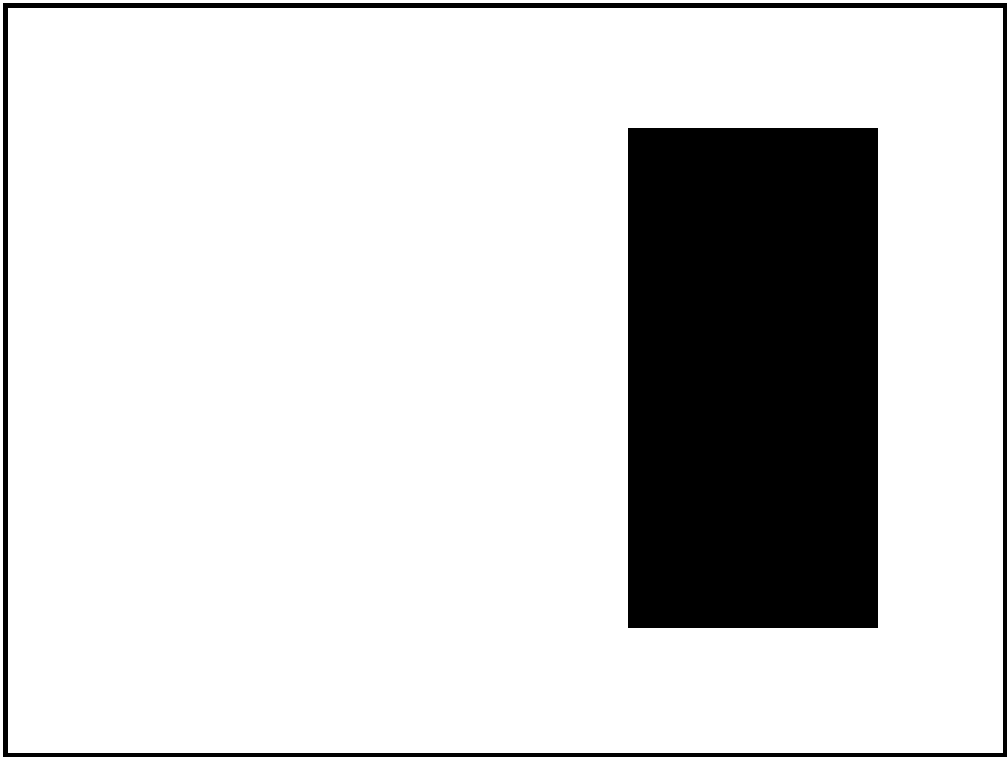} &
            \wimpanel{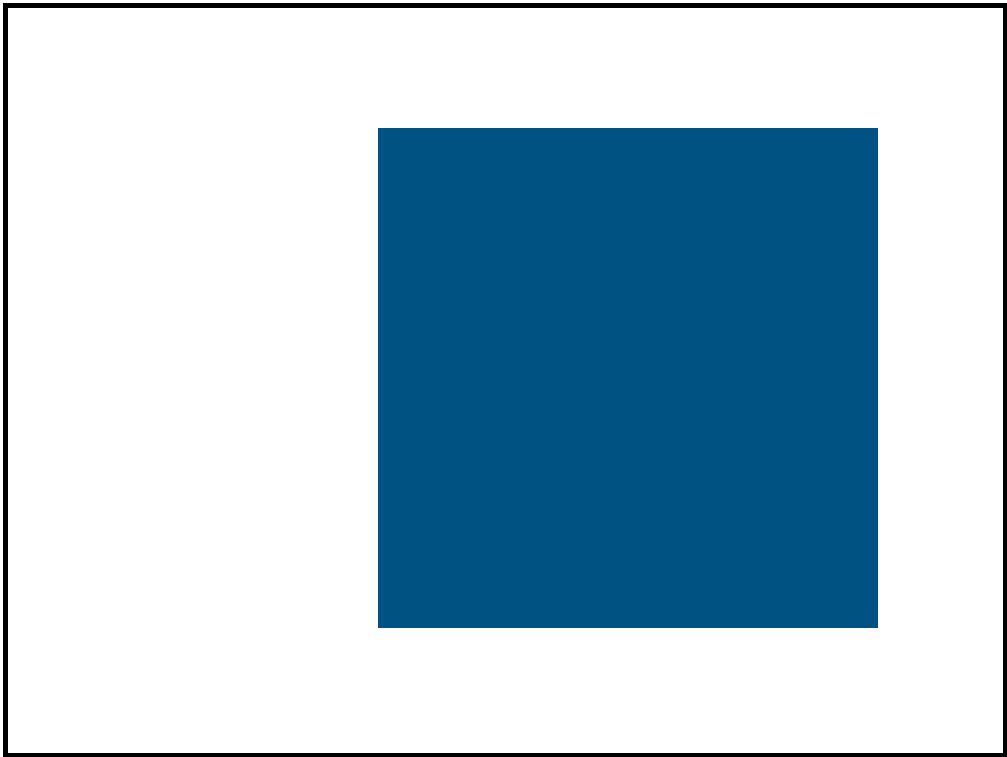} &
            \wimpanel{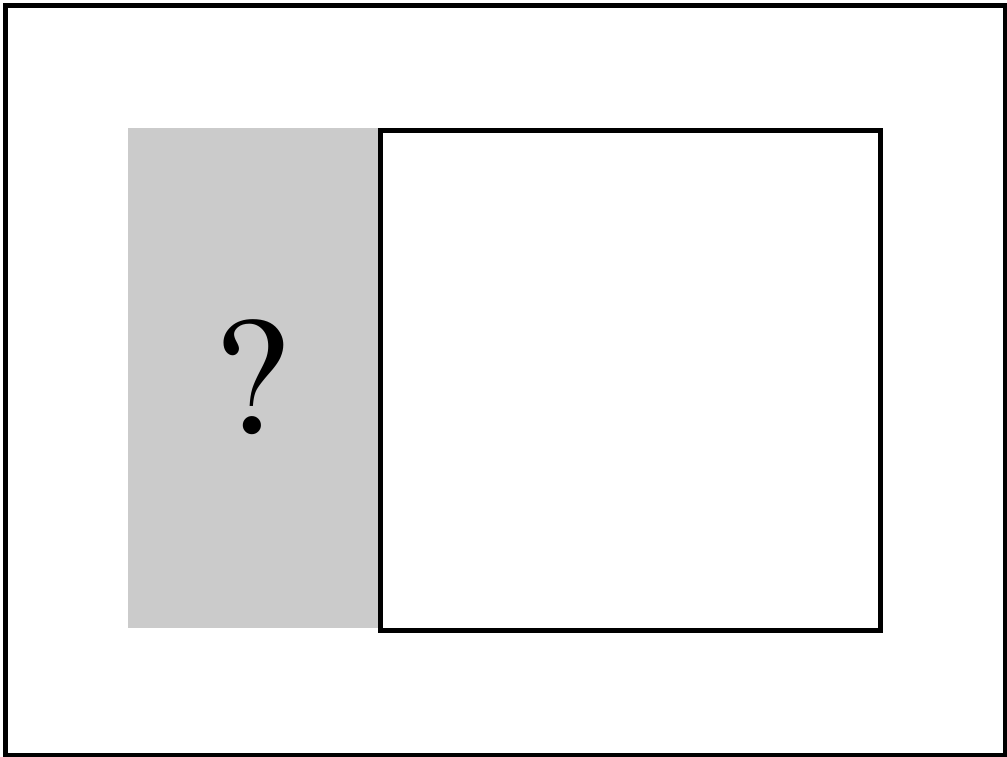} &
            \wimpanel{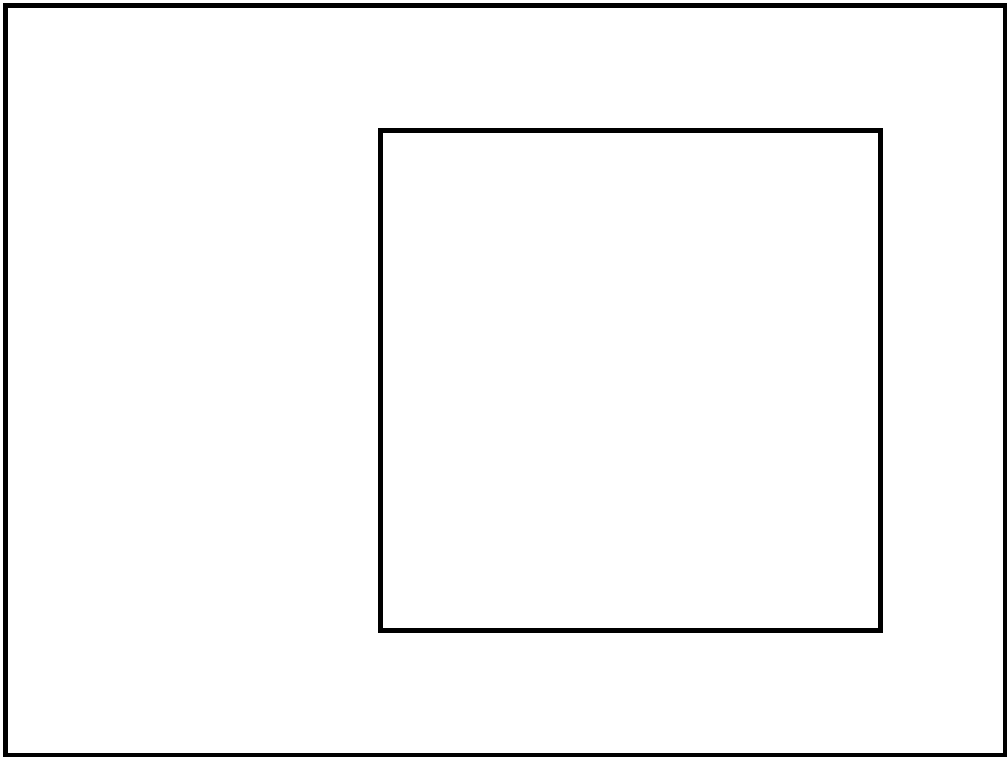} &
            \wimpanel{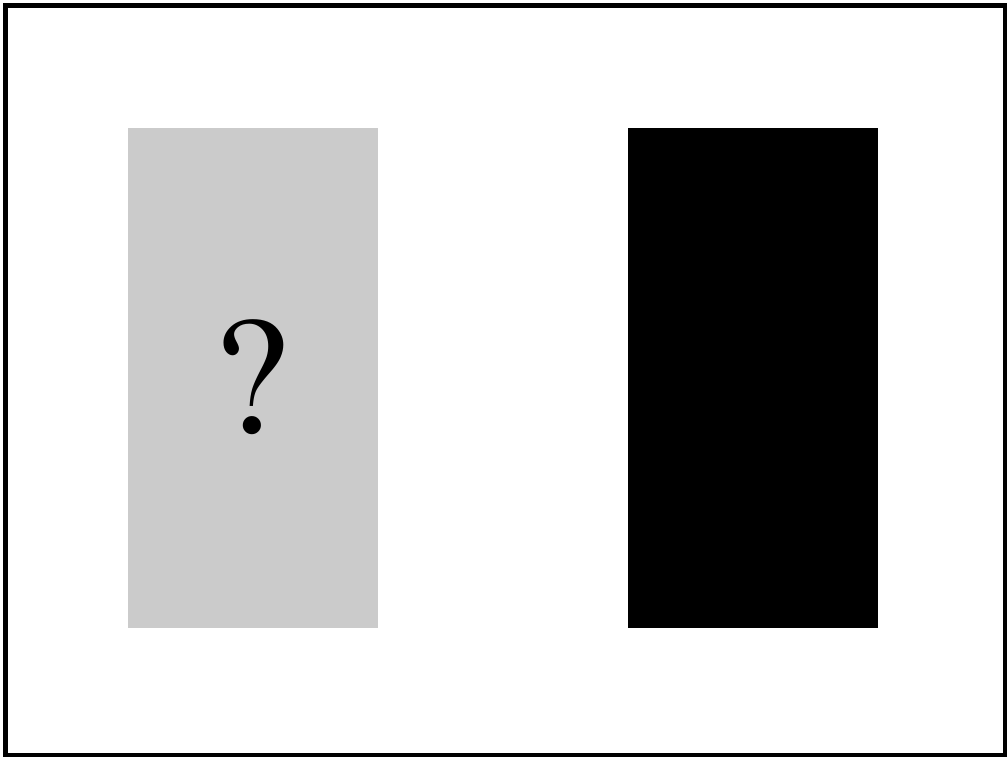} &
            \wimpanel{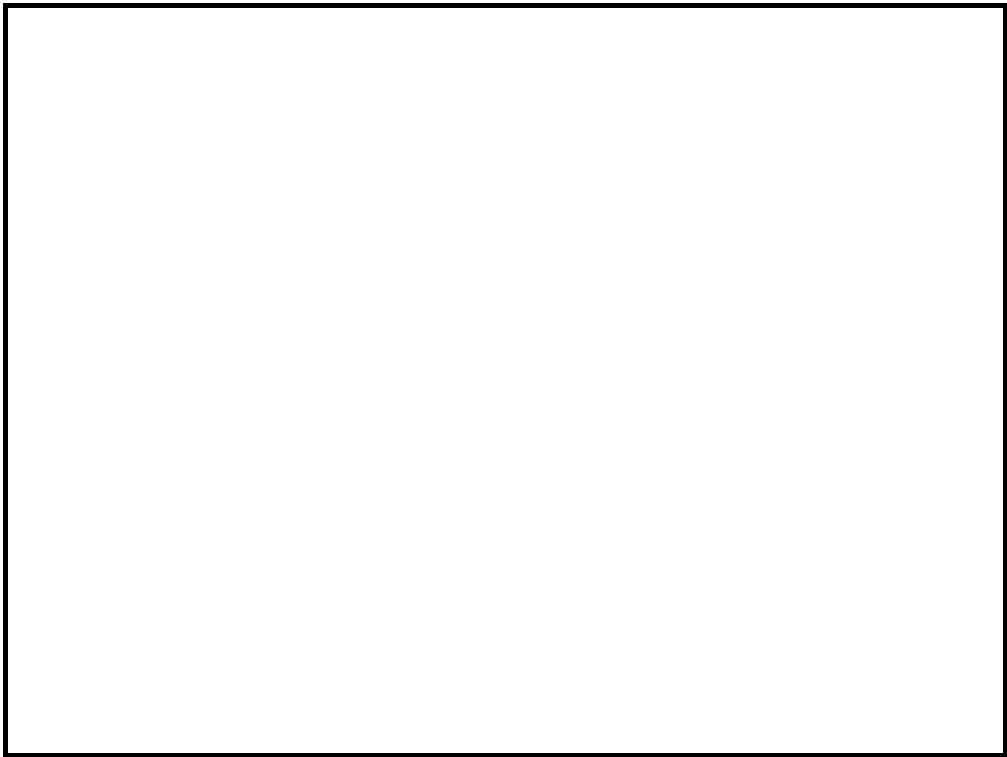} \\
            $I_1$ &
            $M_1$ &
            $O_1$ &
            $I_2$ &
            $D(M_1)$ &
            $R(M_1)$ &
            $D(O_1)$ &
            $R(O_1)$ 
        \end{tabular}
    \end{center}
    \caption{Image frames $I_1$ and $I_2$ show a blue square translating to the right on a static white background. The MB map $M_1$ and Occ map $O_1$ in frame 1 can be warped with direct ($D(\cdot)$) or reverse ($R(\cdot)$) flows to frame 2. The gray rectangles with question marks in $D(M_1)$ and $D(O_1)$ denote regions for which no map values are available because the flow $F_{1\rightarrow 2}$ maps no points into those regions.}
    \label{fig:mb_occ_ex}
\end{figure}

All the previous MB~\cite{eccv18_mb}, Occ~\cite{eccv18_mb, refine_of_occ}, and flow~\cite{eccv18_mb, refine_of_occ, PWC, flownet, flownet2, raft} estimators that use cost volumes use reverse warping (equation \ref{eq:reverse_warp}). We are the first to direct warp for motion analysis. As we  illustrate in Figure \ref{fig:mb_occ_ex} for a $1\rightarrow 2$ mapping, direct warping preserves Occ information \emph{and} provides additional Occ information through the regions that it leaves undefined. Here and elsewhere, we let $M_a$ be the MB maps in frame $a$, and $D(M_a)$ and $R(M_a)$ be the maps in frame $b$ obtained by direct (equation \ref{eq:direct_warp}) and reverse (equation \ref{eq:reverse_warp}) warping of $M_a$. We define $O_a$, $D(O_a)$, and $R(O_a)$ similarly. First, the Occ in $O_1$ (black rectangle) is correctly mapped to the second frame in $D(O_1)$ (recall that the background does not move). In contrast, $R(O_1)$ overwrites this information with the no-Occ information from the foreground square, since $F_{2\rightarrow 1}$ is defined everywhere on that square. Thus, direct warping preserves Occs but reverse mapping does not. Second, $D(O_1)$ has no values in the gray-shaded rectangle with the question mark, which is the part of the background that has become newly visible in frame 2. None of the points that are visible in frame 1 move to that region, and $D(O_1)$ is therefore undefined there. In contrast, $R(O_1)$ uses $F_{2\rightarrow 1}$, which is defined everywhere in frame 2, and $R(O_1)$ is therefore defined everywhere there as well. The presence of undefined values in $D(O_1)$ is a useful source of information for the detection of Occs in the $2\rightarrow 1$ direction, and this information is unavailable when reverse warping is used. Similar considerations hold for the warped MB maps $D(M_1)$ and $R(M_1)$.

\subsection{Motion Boundaries Align with Occlusion Regions}

In most cases, Occs occur near flow field discontinuities: An object in the foreground moves differently from its background, and the resulting curve of flow discontinuity sweeps over the background to make an Occ. Thus, MBs and Occs align, as Figure \ref{fig:occ_grad_mb_align} illustrates.

Some MBs do cut across Occs. For instance, the right edge of the image adds a third MB to the two MBs between the boxes. Also, Occ patterns for thin regions are complex (small potted plant on the left). Furthermore, when the motion difference between foreground and background is parallel to the MB, there is no occlusion or dis-occlusion (top of the brown cylinder near the image center). Finally, the ``ground truth'' used in Figure \ref{fig:occ_grad_mb_align} is not perfect: Some MBs are two pixels thick, when they should ideally have measure zero. Of the two pixels at some point on an MB, one is in the foreground and the other is in the background, so they are warped by different flow values. A $D(M_2)$ (blue) MB pixel that is mistakenly warped by the background motion ends up overlapping an $M_1$ (red) pixel, giving purple.

Nonetheless, in spite of exceptions and imperfections, the pattern is clear: MBs are adjacent to Occ boundaries. This pattern is also borne by statistics: We found that $80.5\%$ of MB pixels in MPI-Sintel are within one pixel of an Occ boundary, and $89.3\%$ are within three pixels. MONet therefore decodes MB and Occ features jointly in a cascade of \emph{levels} that go fine-to-coarse. (As discussed later on, we found fine-to-coarse to work better than coarse-to-fine.) The magnitudes of the gradients of the Occ maps at one level are sampled and passed to the next-level MB predictor. Conversely, the MBs at one level are sampled and passed to the next level Occ predictor. MONet also incorporates an explicit attention mechanism~\cite{Zhang_2018_CVPR} to focus the MB predictor on boundaries of Occs. Specifically, an attention map $A\in [0, 1]^{w\times h}$ is constructed at each level from the gradient of the appropriate Occ map at that level. This map is then is multiplied with the MB feature at that level with the Hadamard product before it is sub-sampled and passed to the subsequent level (see Figure \ref{fig:cb_a_maps} (c)).

\setlength{\tabcolsep}{10pt} 

\subsection{Occlusions or Motion Boundaries Have High Residual Cost}
\begin{figure}[!]
    \centering
    \begin{tabular}{cccc}
        \includegraphics[width=0.24\linewidth]{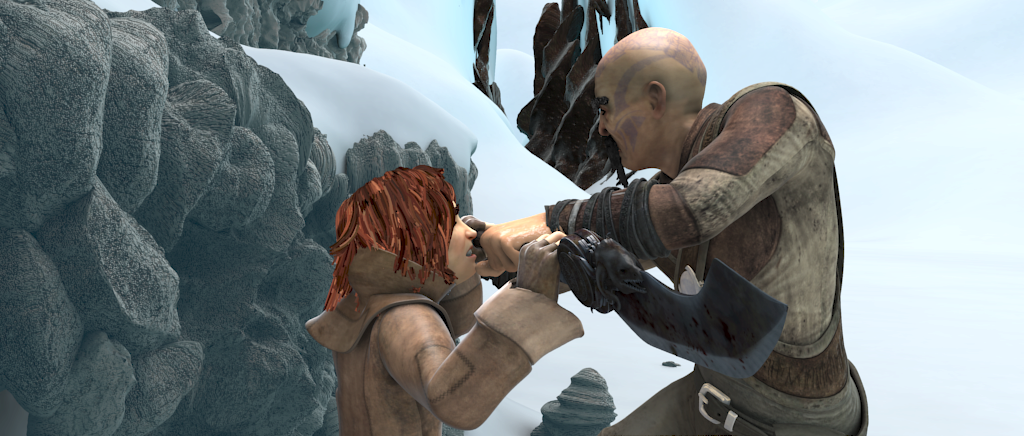} & 
        \includegraphics[width=0.24\linewidth]{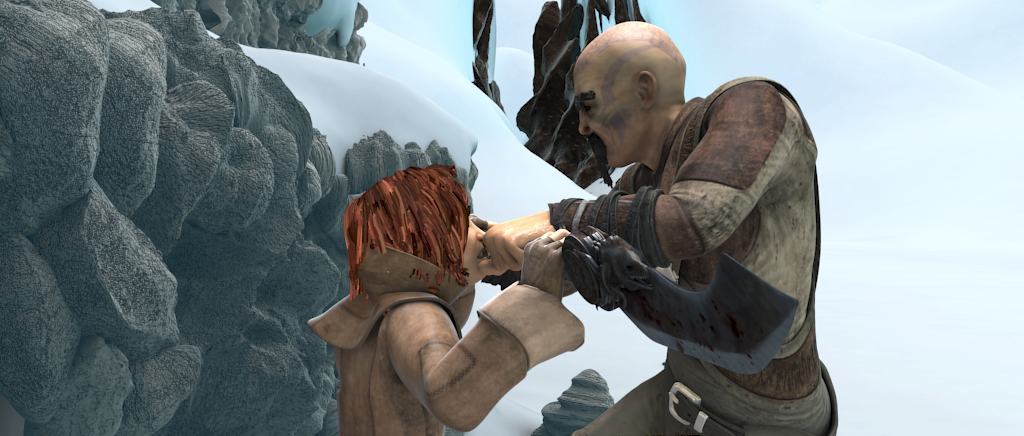} & 
        \includegraphics[width=0.24\linewidth]{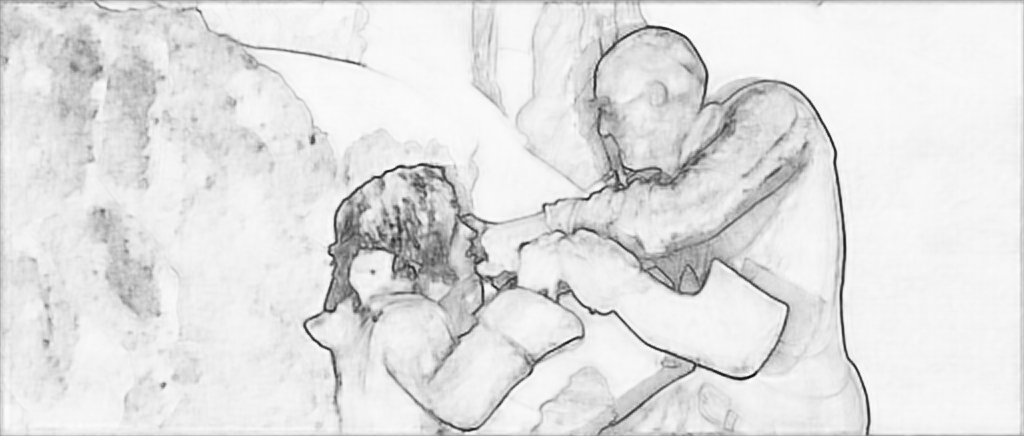} & 
        \includegraphics[width=0.24\linewidth]{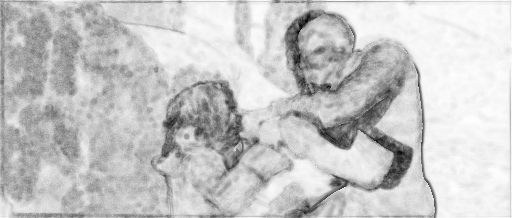} \\
        (a) $I_1$ & (b) $I_2$ & (c) $A_1$ & (d) $B_1$ \\ 
    \end{tabular}
    \caption{Attention map $A_1$ (c) and cost block $B_1$ (d) computed in the finest resolution, using an example from Sintel~\cite{mpi} (a and b). Black is large and white is small for $A_1$ and $B_1$.
    }
    \label{fig:cb_a_maps}
\end{figure}

The discrepancies between features computed from the two frames also provide evidence
for Occs and MBs. Specifically, suppose that a feature vector $\bff_a$ with receptive field $R_a$ centered at pixel $\bx_a$ in frame $a$ is tentatively matched to a feature vector $\bff_b$ with receptive field $R_b$ centered at $\bx_b$ in frame $b$. Then, if $\bx_a$ and $\bx_b$ do not correspond to each other because of occlusion, the vectors $\bff_a$ and $\bff_b$ are likely to be different from each other. In addition, if, say, $\bx_a$ is on an MB, then $R_a$ contains both foreground and background pixels. Pixels in these two subsets move differently, so it is unlikely that all of them match up with corresponding pixels in $R_b$, again leading to differences between $\bff_a$ and $\bff_b$.
To exploit this intuition, we also construct optional \emph{cost blocks} to measure feature discrepancies. Many SOTA Occ and MB methods~\cite{refine_of_occ,eccv18_mb} construct a four-dimensional \textit{cost volume} $V_1(\bx_1, \bd) \in \mathbb{R}^{w \times h \times (2s+1) \times (2s+1)}$ of the Euclidean distances between a feature at $\mathbf{x_1}$ in the first frame and that of a point at $\bx_1 + \bd$ for displacement $\bd$ within $s$ pixels in each dimension in the second frame. These cost volumes are overkill for us, because we already have flow estimates inputs ($F_{a\rightarrow b}$). Instead, we construct two smaller two-dimensional \textit{cost blocks} $B_1, B_2\in \mathbb{R}^{w\times h}$ where
\begin{equation}
    B_a(\bx_a) = \min_{\bd\in [-s, s]^2} V_a(\bx_a, \bd +F_{a\rightarrow 3-a}(\bx_a)) \;\;\text{for}\;\; a = 1, 2\;.
\end{equation}
High values in either $B_1$ or $B_2$ often signal vicinity to MBs or Occs (see Figure \ref{fig:cb_a_maps} (d)).

\setlength{\tabcolsep}{1pt} 

\section{MONet Architecture and Training Details}\label{sec:architecture}

\tightparagraph{Encoder}
We use a fully-convolutional neural network to learn feature maps at multiple scales from fine to coarse, and compute a cost block on these features at each scale. At each scale, four convolutional layers with leaky ReLU activations~\cite{lrelu} process down-scaled input images and output 32-channel feature maps. Each feature map is down-sampled with stride-2 convolutions and is passed to the next coarser scale.

\tightparagraph{Motion boundary and occlusion predictors}
The two decoders for MBs and Occs are fully-convolutional and Siamese, and process the same set of inputs in opposite temporal directions and in fine-to-coarse manner, where maps predicted in each scale are used to predict maps in the following coarser scale. Each scale has five convolutional layers with $3\times 3$ kernel and leaky ReLUs~\cite{lrelu}. Feature maps from each block pass through a $1\times 1$ convolution layer to reduce the channel dimension to 1, a de-convolution layer to up-sample the map to the original resolution, and a sigmoid layer to output a prediction. A fusion layer~\cite{hed} takes the concatenation of $L$ multi-scale maps and computes a weighted average of the maps using a $1\times 1$ convolution layer with kernels initialized to $1/L$.

\tightparagraph{Attention modules}
Each attention module has three convolutional layers with $3\times 3$ kernels followed by a sigmoid. Similar to MB and Occ prediction maps, the output attention feature maps are passed through a $1\times 1$ convolution layer, a de-convolution layer, and a sigmoid layer to output the full-resolution attention maps. 
\tightparagraph{Fine-to-coarse decoders}
MONet's decoders are Fine-to-Coarse (F2C) (Figure \ref{fig:cnn_architectures} (b)), in contrast with the Coarse-to-Fine (C2F) decoders (Figure \ref{fig:cnn_architectures} (a))
used in current encoder-decoder CNNs that involve cost volume computations~\cite{flownet,refine_of_occ,eccv18_mb,cv_of_occ,PWC}. Skip connections feed the encoder feature maps (horizontal arrows) to decoder layers of corresponding resolution (right pyramid). The only differences between the two types of architecture are (i) replacing every up-sampling layer in the decoder with a down-sampling layer, and (ii) connecting every layer to the immediately \emph{finer} layer below it, rather than the coarser one above. A fusion layer combines the multi-scale predictions into the high-resolution output. Without adding computation, the new F2C architecture improves over C2F (Section \ref{results}).

\tightparagraph{Training}
We minimize the focal loss~\cite{focalLoss} on MB, Occ, and attention maps with Adam~\cite{adam} and with an initial learning rate of $10^{-4}$. We use flow estimated from PWC-Net~\cite{PWC} as input for training, and various flow estimators~\cite{PWC,eccv18_mb,refine_of_occ,raft} for evaluation. We implement MONet in Tensorflow~\cite{TF}, and is available at \href{https://github.com/hannahhalin/MONet}{https://github.com/hannahhalin/MONet}.

\section{Datasets and Performance Evaluation} \label{datasEvals}

We train on FlyingThings3D~\cite{flythings3D} dataset and evaluate on FlyingChairsOcc~\cite{flownet, refine_of_occ} and MPI-Sintel~\cite{mpi} datasets without any fine-tuning. \textbf{FlyingThings3D}~\cite{flythings3D} (FT3D) is created by moving graphics-generated objects along random 3D trajectories and includes 21818 training images and 4248 testing images. \textbf{FlyingChairs}~\cite{flownet} is a synthetic dataset generated by applying random affine transformation to Flickr images as backgrounds, and a set of rendered moving 3D chairs as foreground. It consists of 22872 image pairs and corresponding ground truth flow. \textbf{FlyingChairsOcc}~\cite{refine_of_occ} adds ground-truth forward and backward Occ maps to FlyingChairs. \textbf{MPI-Sintel}~\cite{mpi} contains 23 high resolution sequences of 20 to 50 frames each from the open-source computer-animation short "Sintel". Fast motion and large Occs make this dataset challenging.
We follow the literature~\cite{eccv18_mb,refine_of_occ, LDMB} and evaluate Occ predictions by average $F_1$-score after thresholding the map at 0.5, and evaluate MB predictions by mean Average-Precision (mAP) computed with the BSDS evaluation code~\cite{bsds}.

\section{Results} \label{results}

\begin{figure}[!t]
    \centering  
    \begin{tabular}{cccc}
    \includegraphics[width=0.245\linewidth]{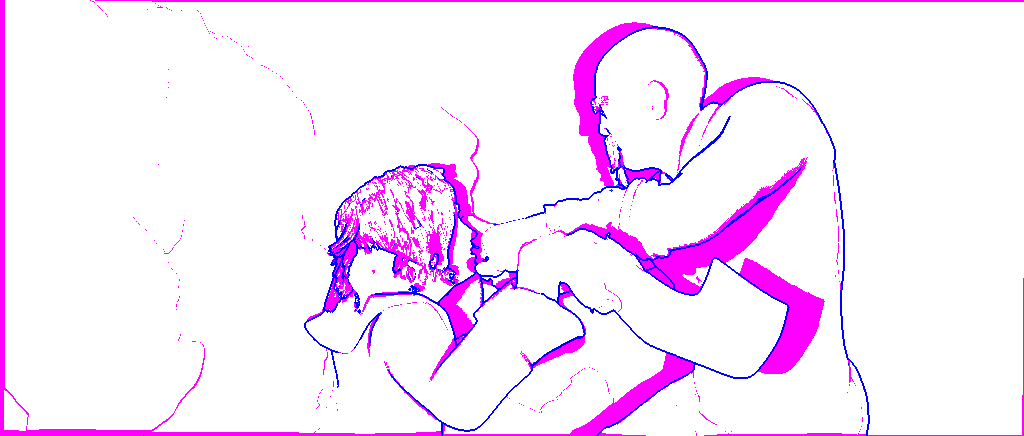} 
    & 
    \includegraphics[width=0.245\linewidth]{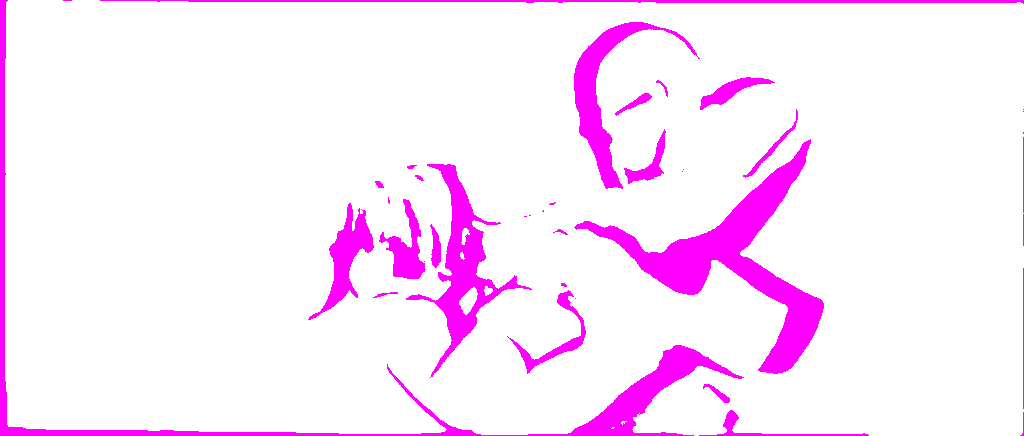} &
    \includegraphics[width=0.245\linewidth]{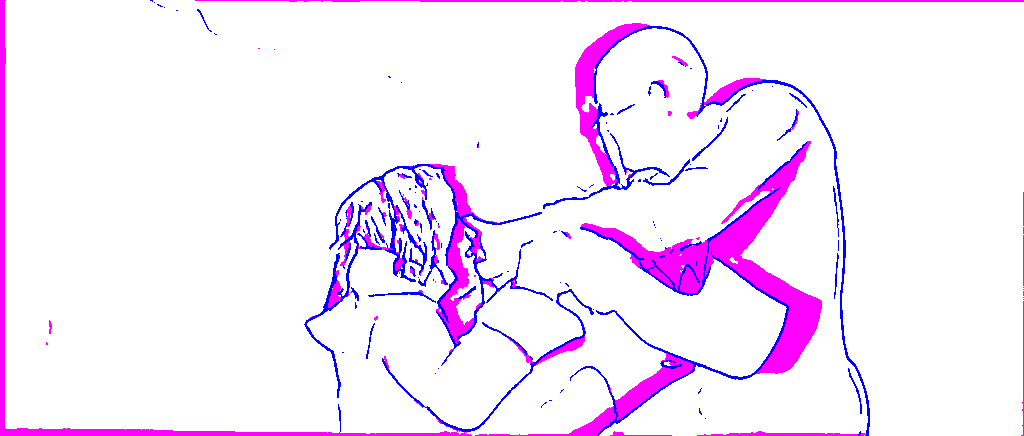}  & 
    \includegraphics[width=0.245\linewidth]{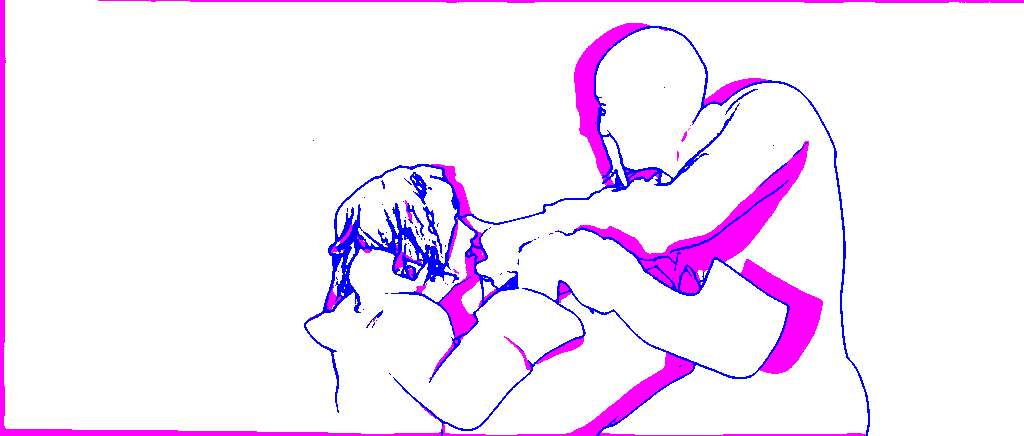}
    \\ 
    \includegraphics[width=0.245\linewidth]{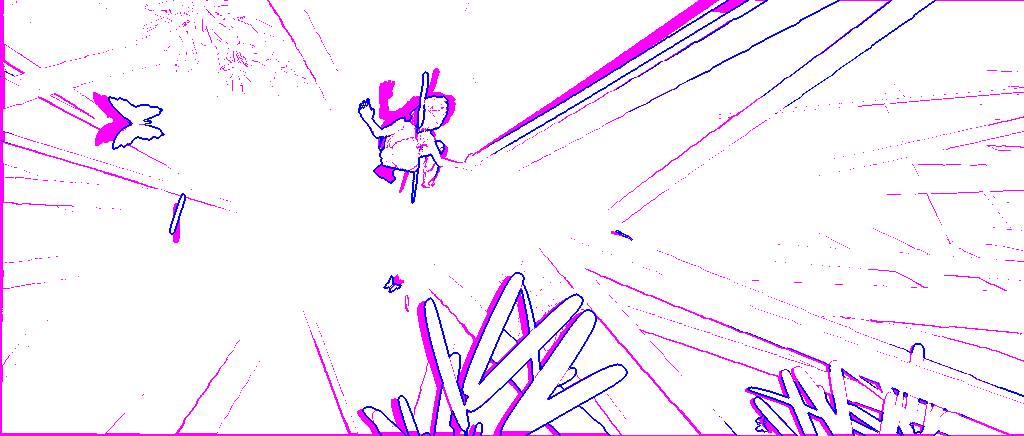} 
    & 
    \includegraphics[width=0.245\linewidth]{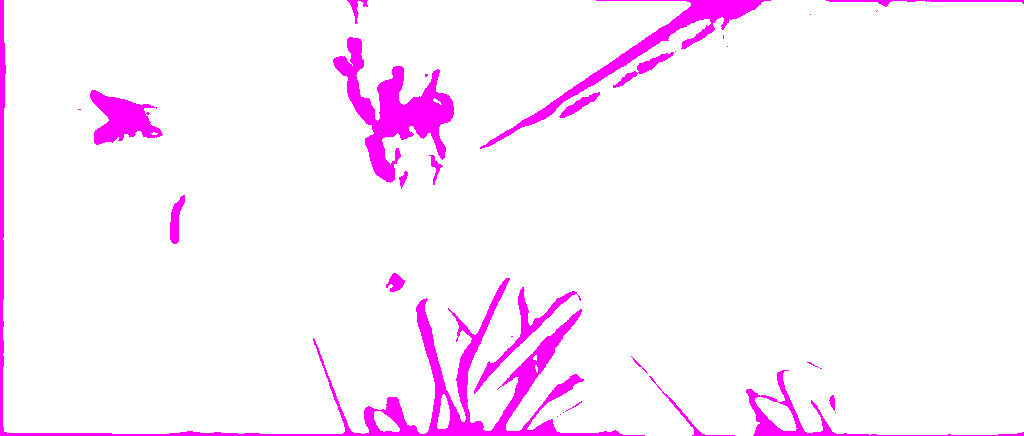} &
    \includegraphics[width=0.245\linewidth]{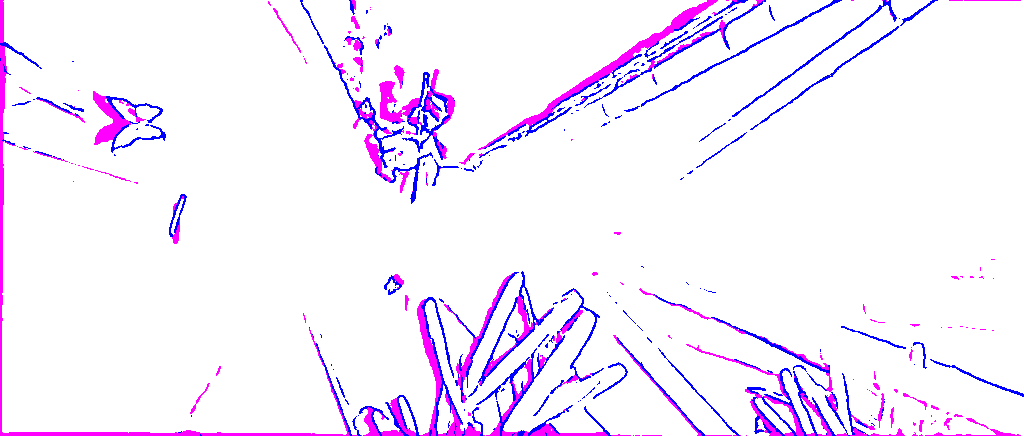} & 
    \includegraphics[width=0.245\linewidth]{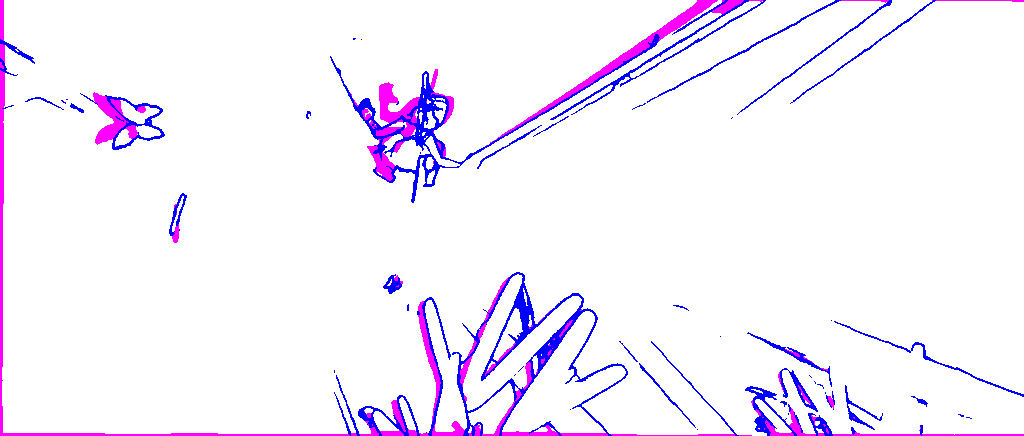}
    \\ 
    \includegraphics[width=0.245\linewidth]{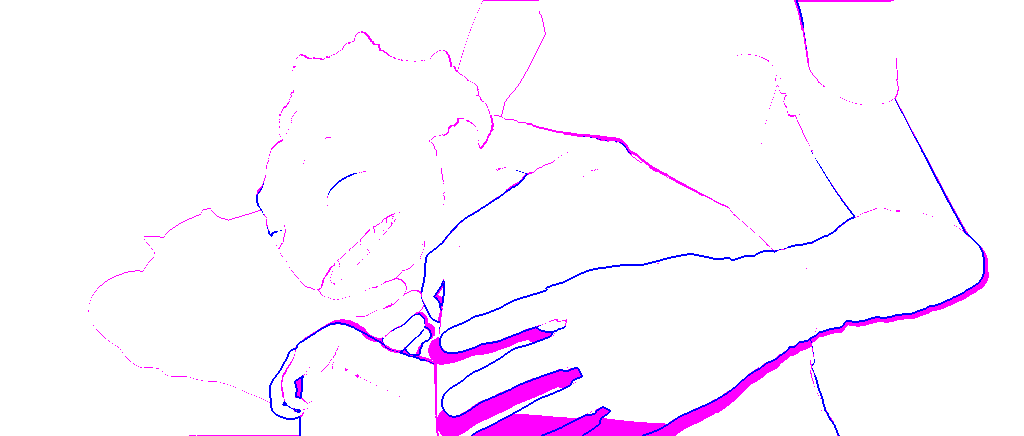} 
    & 
    \includegraphics[width=0.245\linewidth]{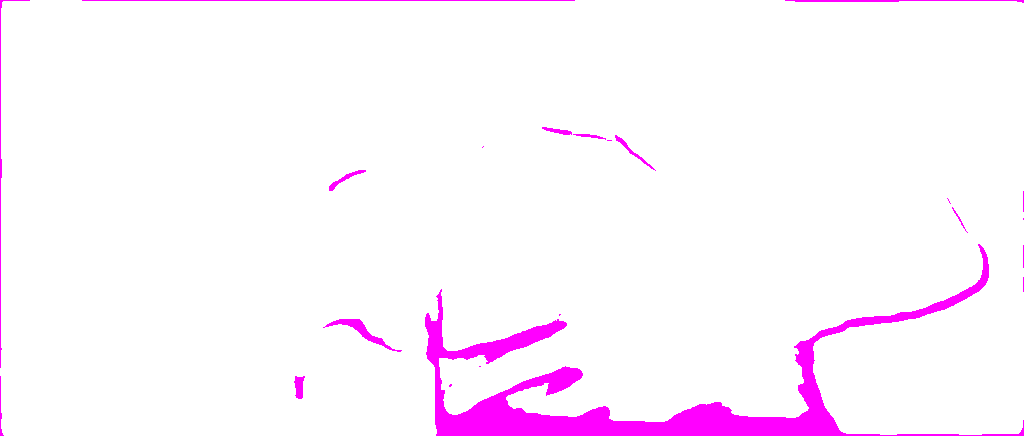} &
    \includegraphics[width=0.245\linewidth]{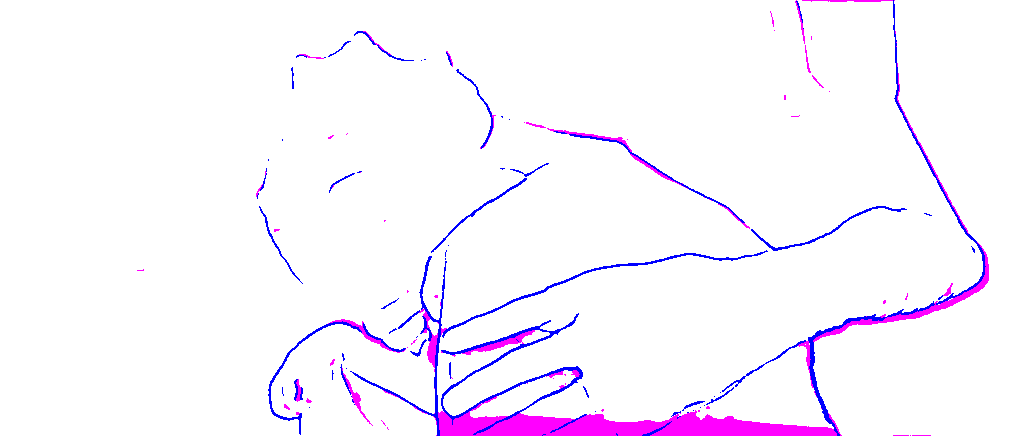} & 
    \includegraphics[width=0.245\linewidth]{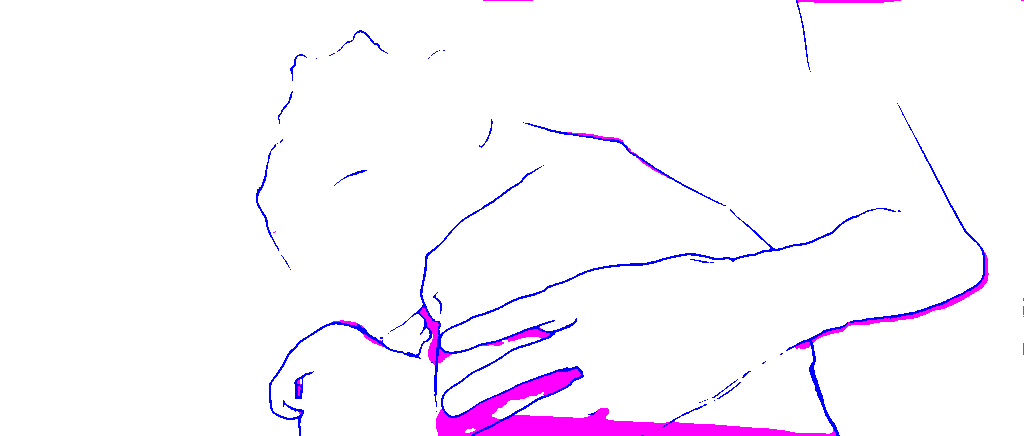}
    \\ 
    Label 
    & \cite{refine_of_occ} & \cite{eccv18_mb}  & MONet
    
    \\
    \end{tabular}
    \caption{MB (blue) and Occ (magenta) predictions of examples from Sintel~\cite{mpi}, thresholded at 0.5. MONet yields precise and clean predictions. (Best viewed magnified and in color.) 
    } 
    \label{fig:visualization_mb_occ_preds}
\end{figure}

MONet outperforms the state of the art in both MB~\cite{eccv18_mb} and Occ~\cite{refine_of_occ} detection. Table \ref{tab:sota_performance_occ_mb} compares MONet to the existing SOTA MB detectors, LDMB~\cite{LDMB} and FlowNet-CSS~\cite{eccv18_mb}, and to the SOTA Occ detectors, FlowNet-CSSR-ft-sd~\cite{eccv18_mb} and IRR-PWC~\cite{refine_of_occ}. FlowNet-CSS is trained on FlyingChairs~\cite{flownet} and FT3D~\cite{flythings3D} to achieve the current SOTA performance on MB detection, and it is further fine-tuned on ChairsSDHom~\cite{flownet2} to obtain FlowNet-CSSR-ft-sd for Occ detection. IRR-PWC is trained on FlyingChairsOcc and FT3D, and LDMB~\cite{LDMB} is trained on Sintel. Performance of our proposed MONet is obtained by training on FT3D only \textit{without any fine tuning on Sintel or FlyingChairsOcc}. We also include a baseline MB performance by taking the gradient of estimated flow from RAFT. Specifically, we cap the flow gradients at $25\%$, $50\%$, $75\%$, and $100\%$ of the maximum gradient over the entire dataset, and report the highest performance. MONet bests the SOTA methods for both tasks in all datasets (bold). Figure \ref{fig:visualization_mb_occ_preds} shows precise and clean MB and Occ predictions by MONet.

\setlength{\tabcolsep}{6pt}
\begin{table}[!t]
    \begin{center}
    \begin{tabular}{c|cccc|ccc}
        & \multicolumn{4}{c|}{Motion Boundary ($mAP$)} & \multicolumn{3}{c}{Occlusion ($F_1$)} \\
        Dataset & Baseline &\cite{LDMB} & \cite{eccv18_mb}
         & MONet & \cite{eccv18_mb} & \cite{refine_of_occ} &
        MONet \\ \midrule
        FlyingChairsOcc~\cite{refine_of_occ} & - & - & - & - & 78.9 & 75.7 & \textbf{82.7} \\ 
        Sintel (Clean)~\cite{mpi} &72.6 & 76.3  & 86.3  &\textbf{93.1} & 70.3 & 71.2 & \textbf{74.4}\\ 
        Sintel (Final)~\cite{mpi} &62.9 & 68.5  & 79.5 &\textbf{79.8}& 65.4 & 66.9 & \textbf{68.7}  \\
    \end{tabular}
    \end{center}
    \caption{Average $F_1$ score for \textbf{occlusion detection} (right) and mean average precision (mAP) for \textbf{motion boundary detection} (left). Our MONet bests the state of the art for both tasks \emph{without any fine tuning on FlyingChairsOcc or Sintel} (bolded).}
    \label{tab:sota_performance_occ_mb}
\end{table}

\setlength{\tabcolsep}{1pt} 

\setlength{\tabcolsep}{0pt} 
\begin{figure}[!t]
    \centering
    \begin{tabular}{cccc}
    \includegraphics[width=0.25\linewidth]{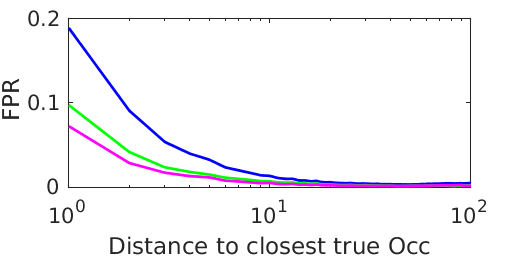} & 
    \includegraphics[width=0.25\linewidth]{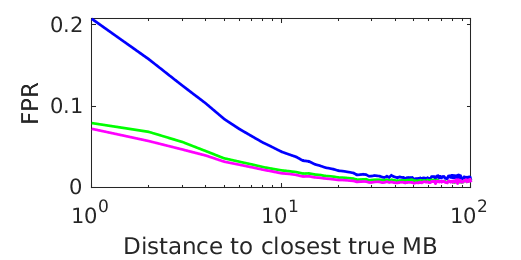} &
    \includegraphics[width=0.25\linewidth]{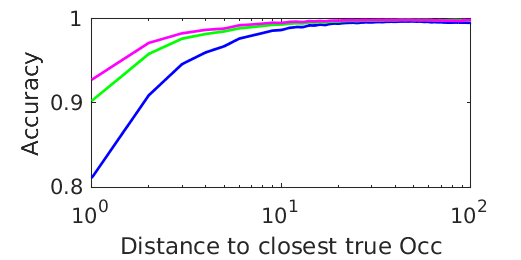} & 
    \includegraphics[width=0.25\linewidth]{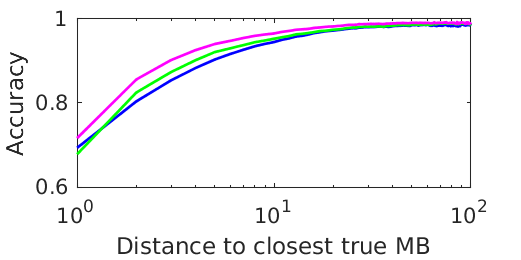} \\ 
    \end{tabular}
    \caption{False Positive Rate (FPR) and accuracy for Occ detection by IRR-PWC~\cite{refine_of_occ} (blue), FlowNet-CSSR-ft-sd~\cite{eccv18_mb} (green), and our MONet (magenta), stratified by the distance to the closest true Occ or MB. Smaller is better for FPR, and larger is better for accuracy.}
    \label{fig:occ_est_perf_stratified}
\end{figure}

Figure \ref{fig:occ_est_perf_stratified} stratifies occlusion performance in terms of false-positive rate and accuracy by each point's distance to the closest Occ or MB. All three methods, IRR-PWC, FlowNet-CSS, and MONet, degrade closer to MBs or Occs. While FlowNet-CSS and MONet detect MBs and Occs jointly, IRR-PWC does not predict MBs and degrades most sharply. MONet shows the best performance across all distances in all plots: Joint detection helps Occ detection, especially when the relationships between MBs and Occs are leveraged explicitly. 

Table \ref{tab:perf_by_component} shows the effects of removing combinations of the proposed components in MONet. Even without all the components, MONet still improves over the SOTA methods~\cite{eccv18_mb,refine_of_occ} (underlined). Using everything yields the best performance in both tasks (bold).

\setlength{\tabcolsep}{15pt} 
\begin{table}[!t]
    \centering
    \begin{tabular}{c|cccccccc}
        & $-DAB$ & $-AB$& $-B$ & $-A$ & MONet \\ \hline
        MB ($mAP$) & \underline{90.8} 
        & \underline{92.1} 
        & \underline{91.6} 
        &\underline{92.2} 
        & \textbf{\underline{93.1}} \\
        Occ ($F_1$) & 69.8 
        &\underline{71.6} &\underline{74.0}& \underline{73.6} & \textbf{\underline{74.4}} \\
    \end{tabular}
    \caption{Effect of proposed components, direct warping ($D$), attention ($A$), and Cost Block ($B$), in Occ and MB detection performance on Sintel. \textit{Even without all the proposed components, MONet still improves over the SOTA methods (underlined)}.}
    \label{tab:perf_by_component}
\end{table}

Table \ref{tab:flowEstInput} evaluates MONet with various input flows~\cite{classicnl,PWC,eccv18_mb,refine_of_occ,raft}.
Performance in both MB and Occ detection increases with better input flow estimates. Even with Classic+NL~\cite{classicnl}, also used by LDMB, MO-Net still outperforms the SOTA for MB detection. Similarly, using the flow from the SOTA MB and Occ estimators~\cite{eccv18_mb,refine_of_occ}, MONet outperforms SOTA for both MB and Occ detection shown in Table \ref{tab:sota_performance_occ_mb}. Regardless of flow input quality, MONet improves on the prior SOTA performance for both tasks (underlined).

\setlength{\tabcolsep}{10pt} 
\begin{table}[!t]
    \centering
    \begin{tabular}{c|cccccc}
    Flow (EPE) & 6.04~\cite{classicnl} & 2.55~\cite{PWC} & 2.08~\cite{eccv18_mb} & 1.88~\cite{refine_of_occ} & 1.43~\cite{raft} \\ \hline 
    MB ($mAP$) & \underline{90.2} & \underline{91.8} & \underline{92.1} & \underline{92.2} & \underline{\textbf{93.1}} \\
    Occ ($F_1$) & 69.2 & \underline{72.3} & \underline{73.0} & \underline{73.4} & \textbf{\underline{74.4}} \\
    \end{tabular}
    \caption{Effect of flow estimation input quality (End-Point Error) in Occ and MB detection performance of MONet on Sintel~\cite{mpi}.  \textit{Regardless of input flow quality, MONet still improves over the SOTA performers in Table \ref{tab:sota_performance_occ_mb} (underlined).}}
    \label{tab:flowEstInput}
\end{table}

\setlength{\tabcolsep}{6pt} 
\begin{table}[!t]
    \centering
    \begin{tabular}{ccc}
    \begin{tabular}{ccc}
        \multirow{2}{*}{MB ($mAP$)}& Single task & Joint task \\ \cline{2-3}
        & 91.6 & \textbf{93.1}
        \\
    \end{tabular} & & 
    \begin{tabular}{ccc}
        \multirow{2}{*}{Occ ($F_1$)}& Single task & Joint task \\ \cline{2-3}
        & 72.5  & \textbf{74.4}
        \\
    \end{tabular}
    \end{tabular}
    \caption{Effect of joint task learning in MB and Occ detection performance on Sintel~\cite{mpi}.} 
    \label{tab:jointLearning}
\end{table}

\setlength{\tabcolsep}{12pt}
\begin{table}[!t]
    \begin{center}
    \begin{tabular}{c|cc|cc|cc}
        \multirow{3}{*}{Dataset} 
        & \multicolumn{2}{c|}{Occ (\cite{refine_of_occ})}  &\multicolumn{2}{c|}{Occ (ours)} &\multicolumn{2}{c}{MB (ours)} \\
        & C2F & F2C & C2F & F2C & C2F & F2C\\ \midrule
        FlyingChairsOcc & 75.7 & \textbf{78.5} &80.9 & \textbf{82.7}&- &-\\ 
        Sintel (Clean) & 71.2 & \textbf{71.9} & 73.3 
        & 
        \textbf{74.4}& 91.2 
        & \textbf{93.1}\\ 
        Sintel (Final) & 66.9 & \textbf{67.1}  & 65.6
        & \textbf{68.7} &72.4 & \textbf{79.8} \\
    \end{tabular}
    \end{center}
    \caption{Performance of MB ($mAP$) and Occ ($F_1$) detection with C2F and F2C decoders. \textit{The F2C version outperforms the C2F version for both IRR-PWC~\cite{refine_of_occ} and MONet.}}
    \label{tab:exp_results_c2f_f2c}
\end{table}

Table \ref{tab:jointLearning} compares the performance of MONet (joint task solving) with that of estimating MBs or Occs separately. To make a single-task version of MONet, we simply remove the decoder for the other task and all the connections between the two decoders. MONet estimates both MBs and Occs better jointly than separately.

Finally, Table \ref{tab:exp_results_c2f_f2c} compares C2F and F2C decoders. The current SOTA Occ detector, IRR-PWC~\cite{refine_of_occ}, utilizes an encoder-decoder architecture, and we simply reverse the information flow of its decoder to make it F2C, using the same training scheme as for its C2F version. The Table shows that the F2C version of each system consistently outperforms its C2F version. 

We speculate that the F2C predictor preserves spatial details better when compared to C2F predictor as the finer predictions do not evolve from the bottleneck features in the C2F predictor that suppress spatial details. Specifically, in a F2C decoder, layers at finer resolution process information that is closest to the full resolution of the input, and can focus on getting the initial predictions right. The overall picture is captured well by coarse resolution predictions, which can be upsampled to full resolution and combined with good predictions along boundaries from the finer predictions. This process does not work in a C2F predictor, because the finer-resolution predictions are made many layers away from the input, and the only fine-resolution information they get is from the skip connections. The flow of information in F2C decoder is consistent with what is done in the edge detection literature~\cite{hed}.

\section{Conclusion} \label{conclusion}
We propose MONet to jointly detect MBs and Occs in both time directions given two video frames and their estimated bi-directional flow. We direct-warp maps between frames, use an attention mechanism to align MBs and Occs, and provide correspondence information with a cost block within an encoder-decoder architecture with F2C decoders. Fine-to-coarse bests coarse-to-fine both for our architecture and for IRR-PWC. This reversal of information flow can be applied at no cost to any encoder-decoder. MONet improves the SOTA for both MBs and Occs on the Sintel and FlyingChairsOcc without any fine-tuning on either dataset.

\tightparagraph{Acknowledgments:}This material is based upon work supported by the National Science Foundation under Grant No. 1909821 and by an Amazon AWS cloud computing award.
\bibliography{bib}

\begin{thebibliography}{35}
\providecommand{\natexlab}[1]{#1}
\providecommand{\url}[1]{\texttt{#1}}
\expandafter\ifx\csname urlstyle\endcsname\relax
  \providecommand{\doi}[1]{doi: #1}\else
  \providecommand{\doi}{doi: \begingroup \urlstyle{rm}\Url}\fi

\bibitem[Abadi et~al.(2016)Abadi, Barham, Chen, Chen, Davis, Dean, Devin,
  Ghemawat, Irving, Isard, Kudlur, Levenberg, Monga, Moore, Murray, Steiner,
  Tucker, Vasudevan, Warden, Wicke, Yu, and Zheng]{TF}
Martin Abadi, Paul Barham, Jianmin Chen, Zhifeng Chen, Andy Davis, Jeffrey
  Dean, Matthieu Devin, Sanjay Ghemawat, Geoffrey Irving, Michael Isard,
  Manjunath Kudlur, Josh Levenberg, Rajat Monga, Sherry Moore, Derek~G. Murray,
  Benoit Steiner, Paul Tucker, Vijay Vasudevan, Pete Warden, Martin Wicke, Yuan
  Yu, and Xiaoqiang Zheng.
\newblock Tensorflow: A system for large-scale machine learning.
\newblock In \emph{12th USENIX Symposium on Operating Systems Design and
  Implementation (OSDI 16)}, pages 265--283, 2016.
\newblock URL
  \url{https://www.usenix.org/system/files/conference/osdi16/osdi16-abadi.pdf}.

\bibitem[Butler et~al.(2012)Butler, Wulff, Stanley, and Black]{mpi}
Daniel~J. Butler, Jonas Wulff, Garrett~B. Stanley, and Michael~J. Black.
\newblock A naturalistic open source movie for optical flow evaluation.
\newblock In {A. Fitzgibbon et al. (Eds.)}, editor, \emph{European Conference
  on Computer Vision}, Part IV, LNCS 7577, pages 611--625. Springer-Verlag,
  October 2012.

\bibitem[Chen and Koltun(2016)]{fullflow}
Qifeng Chen and Vladlen Koltun.
\newblock Full flow: Optical flow estimation by global optimization over
  regular grids.
\newblock In \emph{2016 {IEEE} Conference on Computer Vision and Pattern
  Recognition, {CVPR} 2016, Las Vegas, NV, USA, June 27-30, 2016}, pages
  4706--4714. {IEEE} Computer Society, 2016.
\newblock \doi{10.1109/CVPR.2016.509}.
\newblock URL \url{https://doi.org/10.1109/CVPR.2016.509}.

\bibitem[Dollár and Zitnick(2013)]{sed}
Piotr Dollár and Lawrence Zitnick.
\newblock Structured forests for fast edge detection.
\newblock In \emph{IEEE International Conference on Computer Vision}, 2013.
\newblock \doi{10.1109/ICCV.2013.231}.

\bibitem[Dosovitskiy et~al.(2015)Dosovitskiy, Fischer, Ilg, H{\"a}usser,
  Haz{\i}rba{\c{s}}, Golkov, v.d. Smagt, Cremers, and Brox]{flownet}
A.~Dosovitskiy, P.~Fischer, E.~Ilg, P.~H{\"a}usser, C.~Haz{\i}rba{\c{s}},
  V.~Golkov, P.~v.d. Smagt, D.~Cremers, and T.~Brox.
\newblock Flownet: Learning optical flow with convolutional networks.
\newblock In \emph{IEEE International Conference on Computer Vision}, 2015.
\newblock URL
  \url{http://lmb.informatik.uni-freiburg.de/Publications/2015/DFIB15}.

\bibitem[Fu et~al.(2016)Fu, Wang, Tao, and Black]{occ4}
Huan Fu, Chaohui Wang, Dacheng Tao, and Michael~J. Black.
\newblock Occlusion boundary detection via deep exploration of context.
\newblock In \emph{The IEEE Conference on Computer Vision and Pattern
  Recognition}, June 2016.

\bibitem[Horn and Schunck(1981)]{of0}
Berthold~K.P. Horn and Brian~G. Schunck.
\newblock Determining optical flow.
\newblock \emph{Artificial Intelligence}, 1981.

\bibitem[Humayun et~al.(2011)Humayun, Aodha, and Brostow]{occ2}
Ahmad Humayun, Oisin~Mac Aodha, and Gabriel~J. Brostow.
\newblock Learning to find occlusion regions.
\newblock \emph{Conference on Computer Vision and Pattern Recognition}, 2011.

\bibitem[Hur and Roth(2019)]{refine_of_occ}
Junhwa Hur and Stefan Roth.
\newblock Iterative residual refinement for joint optical flow and occlusion
  estimation.
\newblock In \emph{IEEE Conference on Computer Vision and Pattern Recognition},
  pages 5747--5756, Long Beach, CA, USA, 2019.

\bibitem[Ilg et~al.(2017)Ilg, Mayer, Saikia, Keuper, Dosovitskiy, and
  Brox]{flownet2}
Eddy Ilg, N.~Mayer, Tonmoy Saikia, Margret Keuper, A.~Dosovitskiy, and T.~Brox.
\newblock Flownet 2.0: Evolution of optical flow estimation with deep networks.
\newblock \emph{2017 IEEE Conference on Computer Vision and Pattern Recognition
  (CVPR)}, pages 1647--1655, 2017.

\bibitem[Ilg et~al.(2018)Ilg, Saikia, Keuper, and Brox]{eccv18_mb}
Eddy Ilg, Tonmoy Saikia, Margret Keuper, and Thomas Brox.
\newblock Occlusions, motion and depth boundaries with a generic network for
  disparity, optical flow or scene flow estimation.
\newblock In \emph{The European Conference on Computer Vision (ECCV)},
  September 2018.

\bibitem[Kamrani et~al.(2019)Kamrani, Naghsh-Nilchi, Sadeghian, Tombari, and
  Navab]{mb_vos}
Zahra Kamrani, Ahmad Naghsh-Nilchi, Hamid Sadeghian, Federico Tombari, and
  Nassir Navab.
\newblock Joint motion boundary detection and cnn-based feature visualization
  for video object segmentation.
\newblock \emph{Neural Computing and Applications}, 2019.
\newblock \doi{10.1007/s00521-019-04448-7}.

\bibitem[Kingma and Ba(2014)]{adam}
Diederik~P. Kingma and Jimmy Ba.
\newblock Adam: A method for stochastic optimization.
\newblock \emph{CoRR}, abs/1412.6980, 2014.
\newblock URL
  \url{http://dblp.uni-trier.de/db/journals/corr/corr1412.html#KingmaB14}.

\bibitem[Kramer(1991)]{kramer1991nonlinear}
Mark~A Kramer.
\newblock Nonlinear principal component analysis using autoassociative neural
  networks.
\newblock \emph{AIChE journal}, 37\penalty0 (2):\penalty0 233--243, 1991.

\bibitem[Lin et~al.(2018)Lin, Goyal, Girshick, He, and Dollar]{focalLoss}
Tsung-Yi Lin, Priyal Goyal, Ross Girshick, Kaiming He, and Piotr Dollar.
\newblock Focal loss for dense object detection.
\newblock \emph{IEEE Transactions on Pattern Analysis and Machine
  Intelligence}, pages 1--1, 07 2018.
\newblock \doi{10.1109/TPAMI.2018.2858826}.

\bibitem[Liu et~al.(2006)Liu, Freeman, and Adelson]{mb0}
Ce~Liu, William~T. Freeman, and Edward~H. Adelson.
\newblock Analysis of contour motions.
\newblock \emph{Advances in Neural Information Processing Systems}, 2006.

\bibitem[Long et~al.(2015)Long, Shelhamer, and Darrell]{fcn}
J.~Long, E.~Shelhamer, and T.~Darrell.
\newblock Fully convolutional networks for semantic segmentation.
\newblock In \emph{2015 IEEE Conference on Computer Vision and Pattern
  Recognition (CVPR)}, pages 3431--3440, Los Alamitos, CA, USA, jun 2015. IEEE
  Computer Society.
\newblock \doi{10.1109/CVPR.2015.7298965}.
\newblock URL
  \url{https://doi.ieeecomputersociety.org/10.1109/CVPR.2015.7298965}.

\bibitem[Maas et~al.(2013)Maas, Hannun, and Ng]{lrelu}
Andrew~L. Maas, Awni~Y. Hannun, and Andrew~Y. Ng.
\newblock Rectifier nonlinearities improve neural network acoustic models.
\newblock In \emph{in International Conference on Machine Learning (ICML)
  Workshop on Deep Learning for Audio, Speech and Language Processing}, 2013.

\bibitem[Martin et~al.(2001)Martin, Fowlkes, Tal, and Malik]{bsds}
David Martin, Charless Fowlkes, Doron Tal, and Jitendra Malik.
\newblock A database of human segmented natural images and its application to
  evaluating segmentation algorithms and measuring ecological statistics.
\newblock In \emph{International Conference on Computer Vision}, volume~2,
  pages 416--423, July 2001.

\bibitem[Mayer et~al.(2016)Mayer, Ilg, H{\"a}usser, Fischer, Cremers,
  Dosovitskiy, and Brox]{flythings3D}
Nikolaus Mayer, Eddy Ilg, Philip H{\"a}usser, Philipp Fischer, Daniel Cremers,
  Alexey Dosovitskiy, and Thomas Brox.
\newblock A large dataset to train convolutional networks for disparity,
  optical flow, and scene flow estimation.
\newblock In \emph{IEEE International Conference on Computer Vision and Pattern
  Recognition}, 2016.
\newblock URL
  \url{http://lmb.informatik.uni-freiburg.de/Publications/2016/MIFDB16}.
\newblock arXiv:1512.02134.

\bibitem[Neoral et~al.(2018)Neoral, Sochman, and Matas]{cv_of_occ}
Michal Neoral, Jan Sochman, and Jiri Matas.
\newblock Continual occlusion and optical flow estimation.
\newblock In C.~V. Jawahar, Hongdong Li, Greg Mori, and Konrad Schindler,
  editors, \emph{Computer Vision - {ACCV} 2018 - 14th Asian Conference on
  Computer Vision, Perth, Australia, December 2-6, 2018, Revised Selected
  Papers, Part {IV}}, volume 11364 of \emph{Lecture Notes in Computer Science},
  pages 159--174. Springer, 2018.
\newblock \doi{10.1007/978-3-030-20870-7\_10}.
\newblock URL \url{https://doi.org/10.1007/978-3-030-20870-7\_10}.

\bibitem[Noh et~al.(2015)Noh, Hong, and Han]{deconvnet}
Hyeonwoo Noh, Seunghoon Hong, and Bohyung Han.
\newblock Learning deconvolution network for semantic segmentation.
\newblock In \emph{2015 IEEE International Conference on Computer Vision
  (ICCV)}, 2015.

\bibitem[Ren et~al.(2005)Ren, Fowlkes, and Malik]{crf}
Xiaofeng Ren, Charless~C. Fowlkes, and Jitendra Malik.
\newblock Scale-invariant contour completion using conditional random fields.
\newblock In \emph{Proc. 10th Int'l. Conf. Computer Vision}, volume~2, pages
  1214--1221, 2005.

\bibitem[Rhemann et~al.(2013)Rhemann, Hosni, Bleyer, Rother, and Gelautz]{cv}
Christoph Rhemann, Asmaa Hosni, Michael Bleyer, Carsten Rother, and Margrit
  Gelautz.
\newblock Fast cost-volume filtering for visual correspondence and beyond.
\newblock \emph{IEEE transactions on Pattern Analysis and Machine
  Intelligence}, 35\penalty0 (2):\penalty0 504--511, February 2013.
\newblock ISSN 0162-8828.
\newblock \doi{10.1109/TPAMI.2012.156}.
\newblock URL \url{http://dx.doi.org/10.1109/TPAMI.2012.156}.

\bibitem[Ronneberger et~al.(2015)Ronneberger, Fischer, and
  Brox]{ronneberger2015u}
Olaf Ronneberger, Philipp Fischer, and Thomas Brox.
\newblock U-net: Convolutional networks for biomedical image segmentation.
\newblock In \emph{International Conference on Medical image computing and
  computer-assisted intervention}, pages 234--241. Springer, 2015.

\bibitem[Schulz and Behnke(2012)]{schulz2012learning}
Hannes Schulz and Sven Behnke.
\newblock Learning object-class segmentation with convolutional neural
  networks.
\newblock In \emph{Proceedings of the European Symposium on Artificial Neural
  Networks (ESANN), April 2012}, 2012.

\bibitem[Shi and Tomasi(1994)]{goodfeatures}
J.~Shi and C.~Tomasi.
\newblock Good features to track.
\newblock In \emph{1994 IEEE Computer Society Conference on Computer Vision and
  Pattern Recognition}, pages 593--600, Seattle, WA, USA, 1994. IEEE Comput.
  Soc. Press.

\bibitem[Spoerri(1991)]{spoerri}
Anselm Spoerri.
\newblock \emph{The Early Detection of Motion Boundaries}.
\newblock PhD thesis, Massachusetts Institute of Technology, Department of
  Brain and Cognitive Sciences, 1991.

\bibitem[Sun et~al.(2014)Sun, Roth, and Black]{classicnl}
Deqing Sun, Stefan Roth, and Michael~J. Black.
\newblock A quantitative analysis of current practices in optical flow
  estimation and the principles behind them.
\newblock \emph{International Journal of Computer Vision (IJCV)}, 106\penalty0
  (2):\penalty0 115--137, 2014.

\bibitem[Sun et~al.(2018)Sun, Yang, Liu, and Kautz]{PWC}
Deqing Sun, Xiaodong Yang, Ming-Yu Liu, and Jan Kautz.
\newblock Pwc-net: Cnns for optical flow using pyramid, warping, and cost
  volume.
\newblock In \emph{Conference on Computer Vision and Pattern Recognition},
  2018.

\bibitem[Teed and Deng(2020)]{raft}
Zachary Teed and Jia Deng.
\newblock Raft: Recurrent all-pairs field transforms for optical flow.
\newblock In \emph{European Conference on Computer Vision}, pages 402--419.
  Springer, 2020.

\bibitem[Weinzaepfel et~al.(2015)Weinzaepfel, Revaud, Harchaoui, and
  Schmid]{LDMB}
Philippe Weinzaepfel, Jerome Revaud, Zaid Harchaoui, and Cordelia Schmid.
\newblock {Learning to Detect Motion Boundaries}.
\newblock In \emph{{Conference on Computer Vision and Pattern Recognition}},
  Boston, United States, June 2015.
\newblock URL \url{https://hal.inria.fr/hal-01142653}.

\bibitem[Xie and Tu(2015)]{hed}
Saining Xie and Zhuowen Tu.
\newblock Holistically-nested edge detection.
\newblock \emph{International Conference on Computer Vision}, 2015.

\bibitem[Xu et~al.(2017)Xu, Ranftl, and Koltun]{dcflow}
Jia Xu, Rene Ranftl, and Vladlen Koltun.
\newblock Accurate optical flow via direct cost volume processing.
\newblock In \emph{IEEE Conference on Computer Vision and Pattern Recognition},
  pages 5807--5815, Honolulu, HI, 2017.

\bibitem[Zhang et~al.(2018)Zhang, Wang, Qi, Lu, and Wang]{Zhang_2018_CVPR}
Xiaoning Zhang, Tiantian Wang, Jinqing Qi, Huchuan Lu, and Gang Wang.
\newblock Progressive attention guided recurrent network for salient object
  detection.
\newblock In \emph{Proceedings of the IEEE Conference on Computer Vision and
  Pattern Recognition (CVPR)}, June 2018.

\end{thebibliography}

\end{document}